\documentclass{bmvc2k}
\usepackage{comment}
\usepackage{wrapfig}
\usepackage{amsmath}
\usepackage{amssymb}
\usepackage{booktabs}
\usepackage{graphicx}
\usepackage{pifont} %for crosses in the table
\usepackage{multirow}
\usepackage{booktabs}
\usepackage{multirow}
\title{\textcolor{white}{space}AVisT: A Benchmark for Visual Object \textcolor{white}{space}Tracking in Adverse Visibility}

\addauthor{\textcolor{white}{space}Mubashir Noman$^{*1}$ \\
            \textcolor{white}{space}Wafa Al Ghallabi$^{*1}$ \\
            \textcolor{white}{space}Daniya  Najiha$^{*1}$  \\
            \textcolor{white}{space}Christoph Mayer$^{*2}$ \\
            \textcolor{white}{space}Akshay Dudhane$^{1}$ \\
            \textcolor{white}{space}Martin Danelljan$^{2}$  \\
            \textcolor{white}{space}Hisham Cholakkal$^{1}$ \\
            \textcolor{white}{space}Salman Khan$^{1, 4}$    \\
            \textcolor{white}{space}Luc Van Gool$^2$  \\
            \textcolor{white}{space}Fahad Shahbaz Khan }{}{1,3}

\addinstitution{MBZUAI}
\addinstitution{CVL, ETH Zurich}
\addinstitution{CVL, Link\"{o}ping University}
\addinstitution{Australian National University}

\runninghead{AViST}{A Benchmark for Visual Object Tracking in Adverse Visibility}

\def\eg{\emph{e.g}\bmvaOneDot}

\begin{document}
\maketitle
\begin{abstract}
    One of the key factors behind the recent success in visual tracking is the availability of dedicated benchmarks. While being greatly benefiting to the tracking research, existing benchmarks do not pose the same difficulty as before with recent trackers achieving higher performance mainly due to (i) the introduction of more sophisticated transformers-based methods and (ii) the lack of diverse scenarios with adverse visibility such as, severe weather conditions, camouflage and imaging effects. 
    
    We introduce AVisT, a dedicated benchmark for visual tracking in diverse scenarios with adverse visibility. AVisT comprises 120 challenging sequences with  80k annotated frames, spanning 18 diverse scenarios broadly grouped into five attributes with 42 object categories.  The  key contribution of AVisT is diverse and challenging scenarios covering severe weather conditions such as, dense fog, heavy rain and sandstorm; obstruction effects including, fire, sun glare and splashing water; adverse imaging effects such as, low-light; target effects including, small targets and distractor objects along with camouflage.  We further benchmark 17 popular and recent trackers on AVisT with detailed analysis of their tracking performance across attributes, demonstrating a big room for improvement in performance. We believe that AVisT can greatly benefit the tracking community by complementing the existing benchmarks, in developing new creative tracking solutions in order to continue pushing the boundaries of the state-of-the-art. Our dataset along with the complete tracking performance evaluation is available at: \url{https://github.com/visionml/pytracking}
    
\end{abstract}

\section{Introduction}
    Visual object tracking is one of the fundamental problems in computer vision, where the objective is to estimate the target state and trajectory in an image sequence, provided only its initial location. The target object is not known a priori and is not constrained to be from a specific object class. Therefore, the main challenge is to accurately learn the appearance of the target object in unconstrained real-world scenarios.

    Recent years have witnessed a significant progress in the field of visual tracking  with a plethora of trackers introduced in the literature. One of the major contributing factors towards these recent advances in tracking is the introduction of several benchmarks~\cite{WU_2015_TPAMI_OTB, Galoogahi_2017_ICCV_NFS, 2018_Muller_Trackingnet, Huang_2021_TPAMI_GOT10k, Fan_2019_CVPR_Lasot}. OTB~\cite{WU_2015_TPAMI_OTB} was one of the first large-scale datasets, containing 100 videos. Afterwards, tracking benchmarks are introduced to evaluate different aspects of tracking such as, the impact of color information~\cite{TempleColor}, fast target motion~\cite{Galoogahi_2017_ICCV_NFS} as well as tracking in aerial imagery~\cite{Mueller_2016_ECCV_UAV123}. More recently, the tracking community has focused on constructing datasets~\cite{2018_Muller_Trackingnet, Fan_2019_CVPR_Lasot,Huang_2021_TPAMI_GOT10k} with large-scale training splits, to benefit from task-specific deep learning. Among these, GOT-10K~\cite{Huang_2021_TPAMI_GOT10k} comprises a large collection of shorter videos, whereas LaSOT~\cite{Fan_2019_CVPR_Lasot} focuses on longer sequences. Moreover, there exist dedicated benchmarks, such as the VOT series~\cite{Kristan_2016_TPAMI_VOT} associated with annual tracking challenge competitions.
\begin{figure}[t]
    \begin{center}
        \includegraphics[width=0.85\linewidth, keepaspectratio]{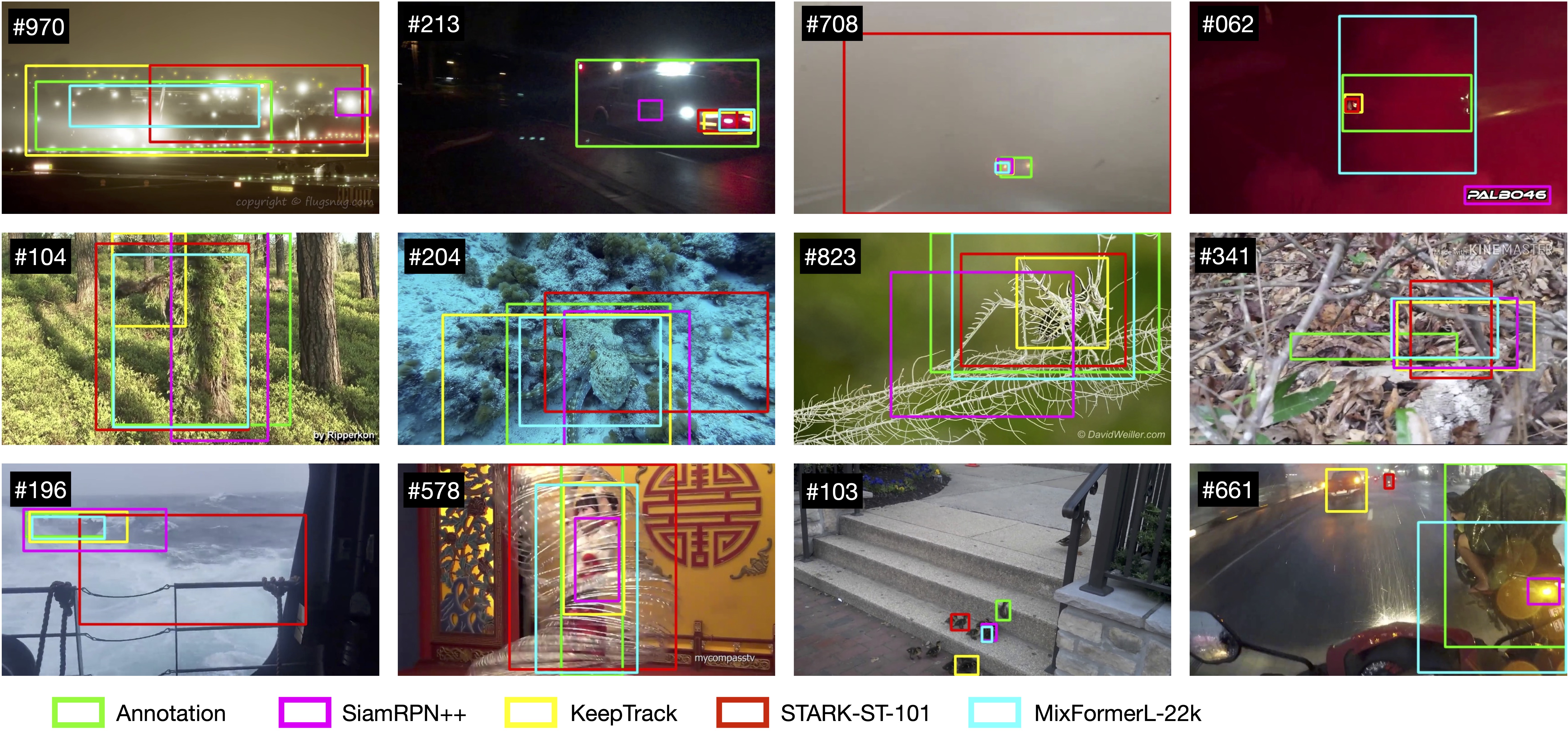}
    \end{center}
    \vspace{-3mm}
    \caption{AVisT comprises challenging diverse tracking scenarios with adverse visibility. The diverse scenarios cover adverse weather conditions, including dense fog, heavy rain and sandstorm; obstruction effects such as, fire and sun glare; illumination effects; target effects, including distractor objects and small targets; along with camouflage. Here, we show individual frames of some representative sequences and visualize with different colors the ground truth annotations and the predicted bounding boxes of four different trackers. The trackers belong to different tracking paradigms: KeepTrack~\cite{Mayer_2021_ICCV_KeepTrack} (Discriminative Classifier), SiamRPN++~\cite{Li_2019_CVPR_SiamRPN++} (Siamese), STARK-ST-101~\cite{Yan_2021_ICCV_STARK} and MixFormerL-22k~\cite{Cui_2022_CVPR_Mixformer} (Transformer).}\label{fig:teaser}
\end{figure}
\begin{table}[t]
	\newcommand{\dist}{\hspace{8pt}}%
	\begin{center}
    	\resizebox{1.0\linewidth}{!}{%
            \begin{tabular}{l@{\dist}c@{\dist}c@{\dist}c@{\dist}c@{\dist}c@{\dist}c@{\dist}}
            	\toprule
            	Datasets            & OTB-100~\cite{WU_2015_TPAMI_OTB}      & UAV123~\cite{Mueller_2016_ECCV_UAV123}        & GOT-10k~\cite{Huang_2021_TPAMI_GOT10k}       & TrackingNet~\cite{2018_Muller_Trackingnet}    & LaSOT~\cite{Fan_2019_CVPR_Lasot}           & AVisT           \\
            	\midrule
                Best Tracker        & TrDiMP \cite{Wang_2021_CVPR_TrDiMP}       & MixFormer-22k \cite{Cui_2022_CVPR_Mixformer}   & MixFormer-1k \cite{Cui_2022_CVPR_Mixformer}  & MixFormerL-22k \cite{Cui_2022_CVPR_Mixformer} & MixFormerL-22k \cite{Cui_2022_CVPR_Mixformer}  & MixFormerL-22k \cite{Cui_2022_CVPR_Mixformer} \\
                Performance         & 71.1         & 70.4            & 71.2          & 83.9           & 70.1            & 56.0            \\
                
            	\bottomrule
            \end{tabular}
    	}
    \end{center}
	\caption{Tracking performance (AUC score) achieved by the top-performing trackers on existing datasets and AVisT. Compared to existing datasets such as, LaSOT and TrackingNet, the performance achieved on AVist is significantly lower highlighting the challenging nature of the proposed dataset.}\label{tab:dataset-comparison}%
\end{table}

While all of these datasets have greatly benefited the tracking research, they no longer pose the same difficulty as before to the current state-of-the-art trackers, due to the rapid progress in the field.
Most notably, the state-of-the-art trackers now achieve AUC scores of above 70\% also on LaSOT (see Tab.~\ref{tab:dataset-comparison}), which is one of the most difficult established datasets. On the other hand, since the introduction of OTB in 2013~\cite{OTB13Ming}, the existence of highly difficult tracking benchmarks has been vital, in order to challenge researchers to designing ever more robust and accurate trackers, applicable for increasingly diverse scenarios. In this work, we therefore set out to develop a new, highly challenging dataset, in order to promote further progress in the visual tracking field. 

We believe that one of the main reasons that the aforementioned datasets do not pose sufficient challenge to new trackers is that diverse scenarios such as, adverse visibility due to weather conditions, camouflage and illumination effects are underrepresented. In practice, robust handling of adverse visibility is essential in many applications. For instance, autonomous driving applications require the target to be tracked under all weather conditions, such as heavy rain, dense fog, and sandstorms. Similarly, rescue missions involving drones require robust and accurate object tracking in adverse scenarios, such as fire, smoke, and strong winds. Further, wildlife conservation often relies on monitoring different animal populations in their natural habitats, where many animal species are difficult to distinguish from the surrounding environments due to their camouflaged appearance.\newline
\noindent\textbf{Contributions:} We propose AVisT, a benchmark for visual object tracking in diverse scenarios with adverse visibility. AVisT better accommodates the difficult conditions encountered in the aforementioned real-world applications, while being severely challenging even for the most recent trackers (see Tab.~\ref{tab:dataset-comparison}). Our dataset comprises 120 challenging sequences, spanning 18 diverse scenarios and 42 object categories. The scenarios cover adverse weather conditions, including heavy rain, dense fog, and hurricane; obstruction effects such as, splashing water, fire, sun glare, and smoke; adverse imaging effects; target effects such as, fast motion and small target; along with camouflage. The proposed AVisT is densely annotated with accurate bounding boxes following a thorough quality control. Moreover, every frame is annotated with flags for occlusion, partial occlusion, out of view, and extreme visibility.

We evaluate 17 popular trackers on AVisT, including the most recent state-of-the-art methods. The best method, MixFormer-22k \cite{Cui_2022_CVPR_Mixformer} which employs  an ImageNet-22K pre-trained backbone achieves an AUC score of only 56.0\%, demonstrating the challenging nature of AVisT. We further analyze the performance of different trackers across attributes, which can provide valuable insights for specific applications. For instance, we note that ImageNet-22K pre-training is important for improved performance on the weather conditions attribute. Fig.~\ref{fig:teaser} shows a qualitative comparison of recent trackers belonging to different tracking paradigms: discriminative classifiers, Siamese networks and transformers.

\section{The AVisT Benchmark}
\subsection{Scenarios and Attributes}

\begin{wrapfigure}{r}{0.46\textwidth}
  \centering
    \includegraphics[width=1\linewidth]{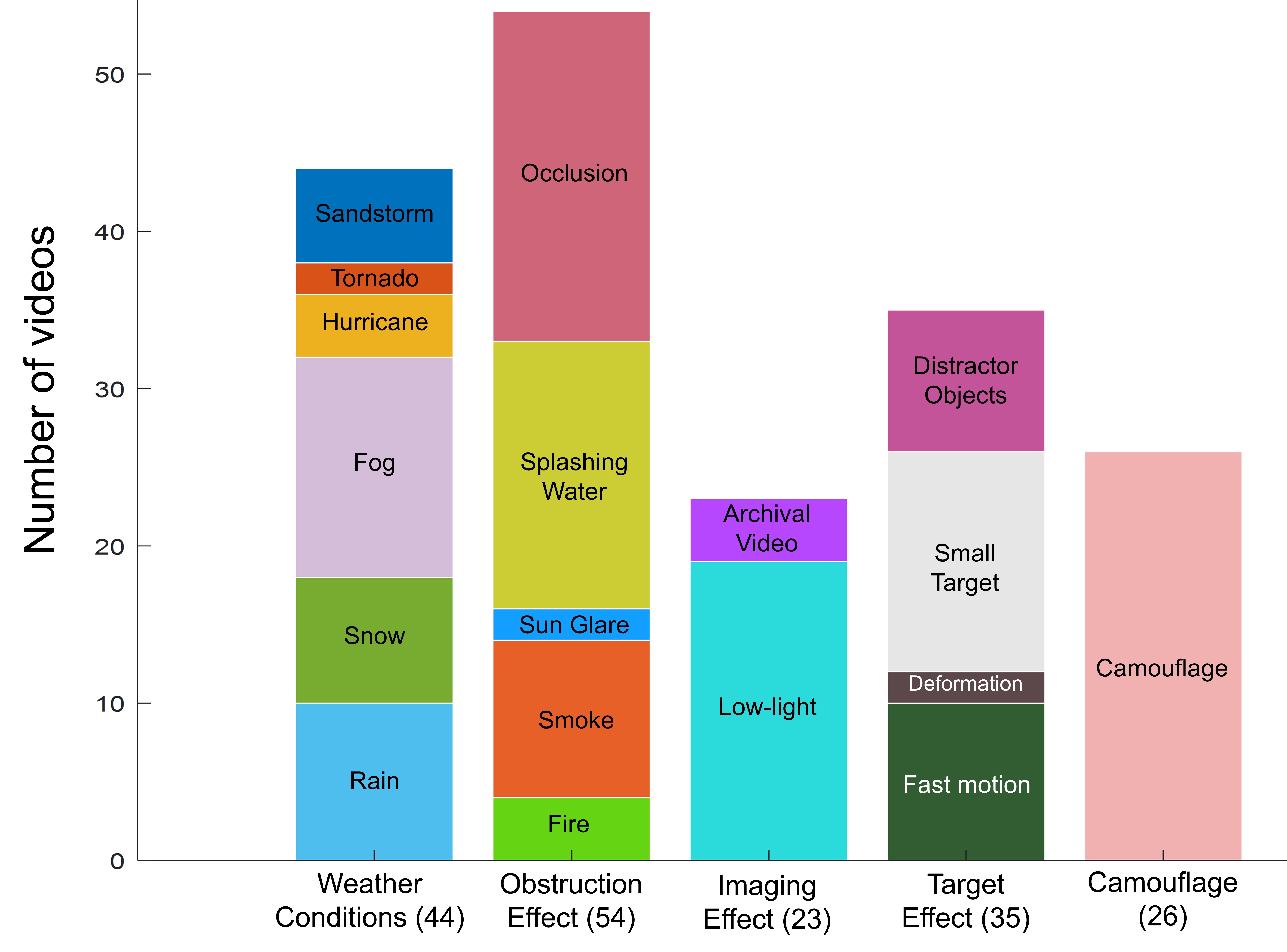} 
  \caption{Distribution of image sequences with respect to scenarios and attributes in the proposed AVisT benchmark. }\label{fig:charts}
\end{wrapfigure}

Our AVisT offers a dedicated dataset that covers a variety of adverse scenarios highly relevant to real-world applications. Importantly, AVisT poses additional challenges to the tracker design due to adverse visibility. To this end, our AVisT covers a wide range of 18 diverse scenarios: rain, fog, hurricane, fire, sun glare, low-light, archival videos, fast motion, distractor objects, occlusion, snow, sandstorm, tornado, smoke, splashing water, camouflage, small objects and deformation. These diverse scenarios are broadly categorized into five attributes: \textit{weather conditions},  \textit{obstruction effects}, \textit{imaging effects}, \textit{target effects} and \textit{camouflage}. A short description of each scenario and their partitioning into attributes are presented in Tab.~\ref{tab:my-table-description}. The frequency of each scenario and attribute is visualized in Fig.~\ref{fig:charts}. Next, we describe the included attributes.

\begin{table}[t!]
\centering
\resizebox{\columnwidth}{!}{%
\begin{tabular}{lll}
\toprule
\textbf{Attribute} & \textbf{Scenario} & \textbf{Description} \\ \midrule
\multirow{6}{*}{Weather Conditions}&Rain            & Heavy rain that compromises the visibility of the target. \\
&Snow           & Heavy snowfall or snow conditions, affecting target visibility. \\
&Fog            & Dense fog that severely affects target visibility.\\
&Sandstorm      & Dense sand and dust in the air, severely impairing target visibility.\\
&Hurricane       & Severe winds accompanied by rain and lightning.\\
&Tornado         & Presence of a tornado, hampering target visibility.\\ \midrule
\multirow{5}{*}{Obstruction Effects}
&Occlusion          & The target is occluded by another object or background structures.\\
&Splashing water    & Splashing water in front of or on the target.\\
&Fire           & Fire obstructing the target as well as causing lighting variation.\\
&Smoke           & Dense smoke obstructing the view of the target.\\
& Sun glare       & Sun glare effects that reduces the visibility of the target.\\ \midrule
\multirow{2}{*}{Imaging Effects}& Low-light          & Poor scene lighting conditions.\\
&Archival          & Monochrome archival videos of poor quality.             \\ \midrule
\multirow{4}{*}{Target Effects}&Fast motion        & The target motion is greater than the target size.                   \\
&Small target      & In at least one frame, the target box is smaller than 500 pixels.    \\
&Distractor objects & Presence of several objects that are visually similar to the target. \\
&Deformation        & The target undergoes shape changes during tracking.                  \\ \midrule
Camouflage&Camouflage         & The target appearance is very similar to the surrounding background.\\ \bottomrule
\end{tabular}}\vspace{6mm}
\caption{A brief description of 18 different adverse scenarios, grouped into five broader attributes (weather conditions, obstruction effects, imaging effects, target effects and camouflage), in the proposed AVisT dataset.}
\label{tab:my-table-description}
\end{table}

\begin{figure}[t]
    \centering
        \scriptsize
        \begin{tabular}{c@{$\;\,$}c@{$\;\,$}c@{$\;\,$}c@{$\;\,$}c@{$\;\,$}c@{$\;\,$}}
            \includegraphics[width=0.13\textwidth]{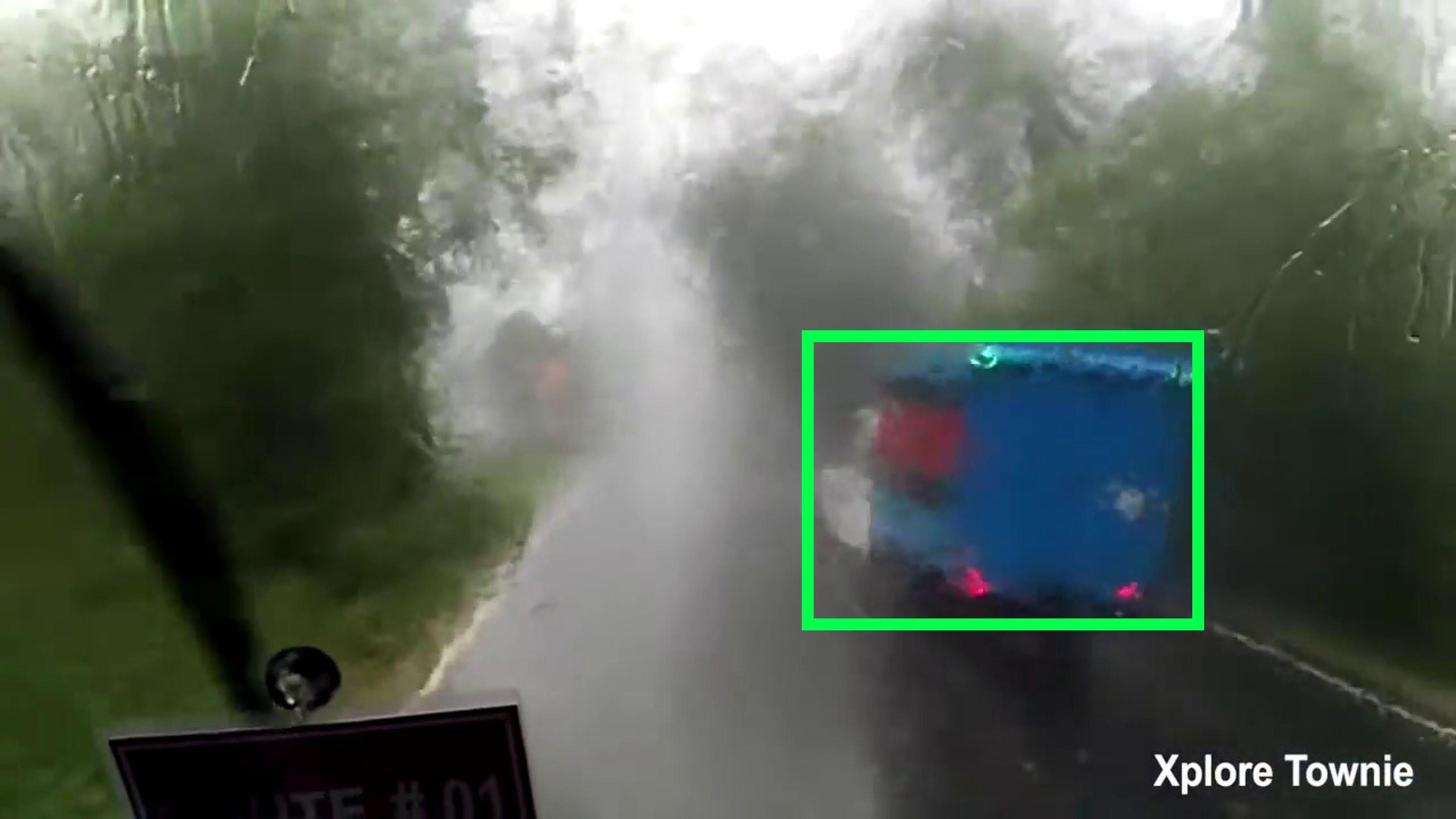}&
            \includegraphics[width=0.13\textwidth]{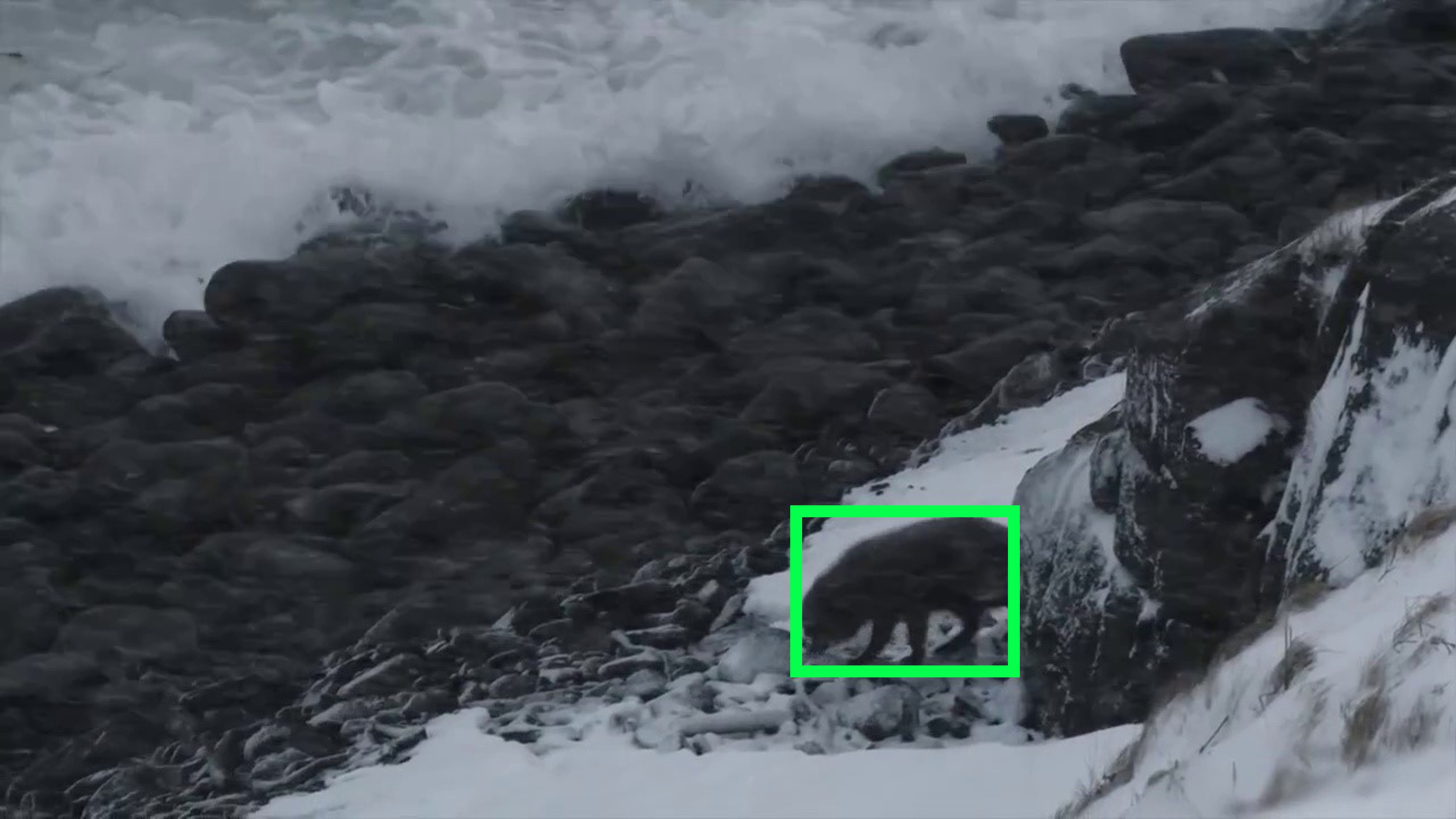}&
            \includegraphics[width=0.13\textwidth]{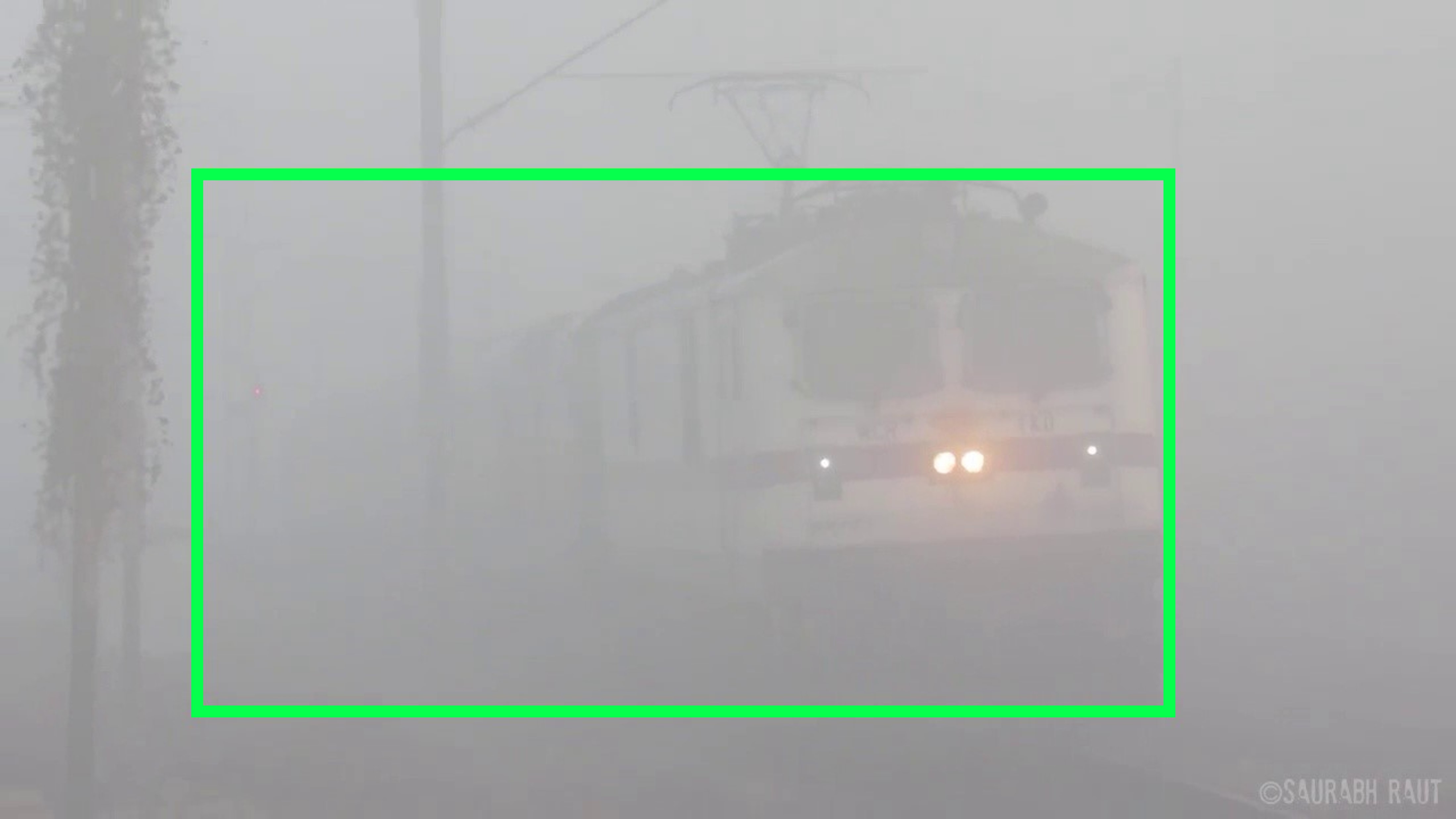}&
            \includegraphics[width=0.13\textwidth]{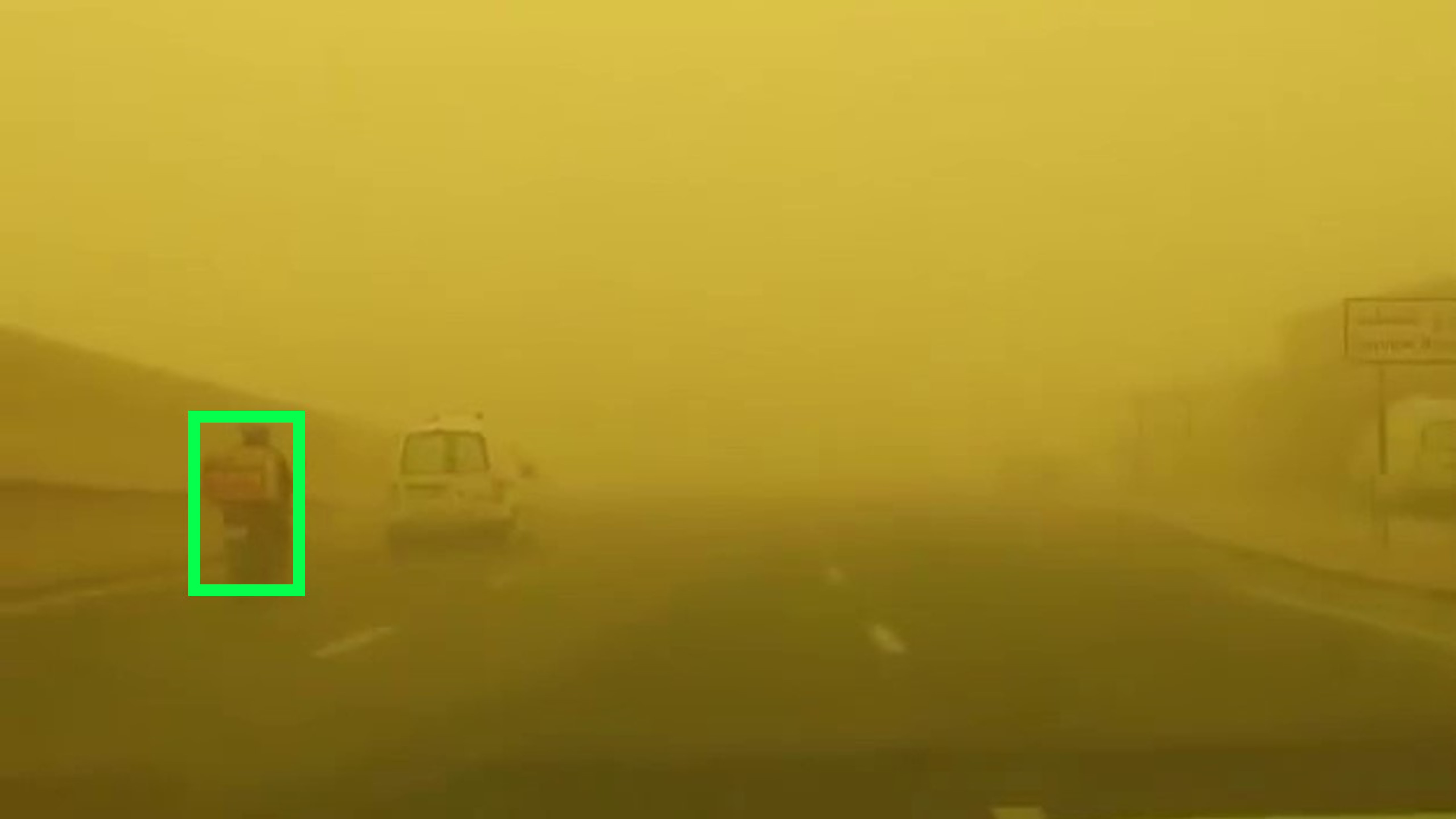}&
            \includegraphics[width=0.13\textwidth]{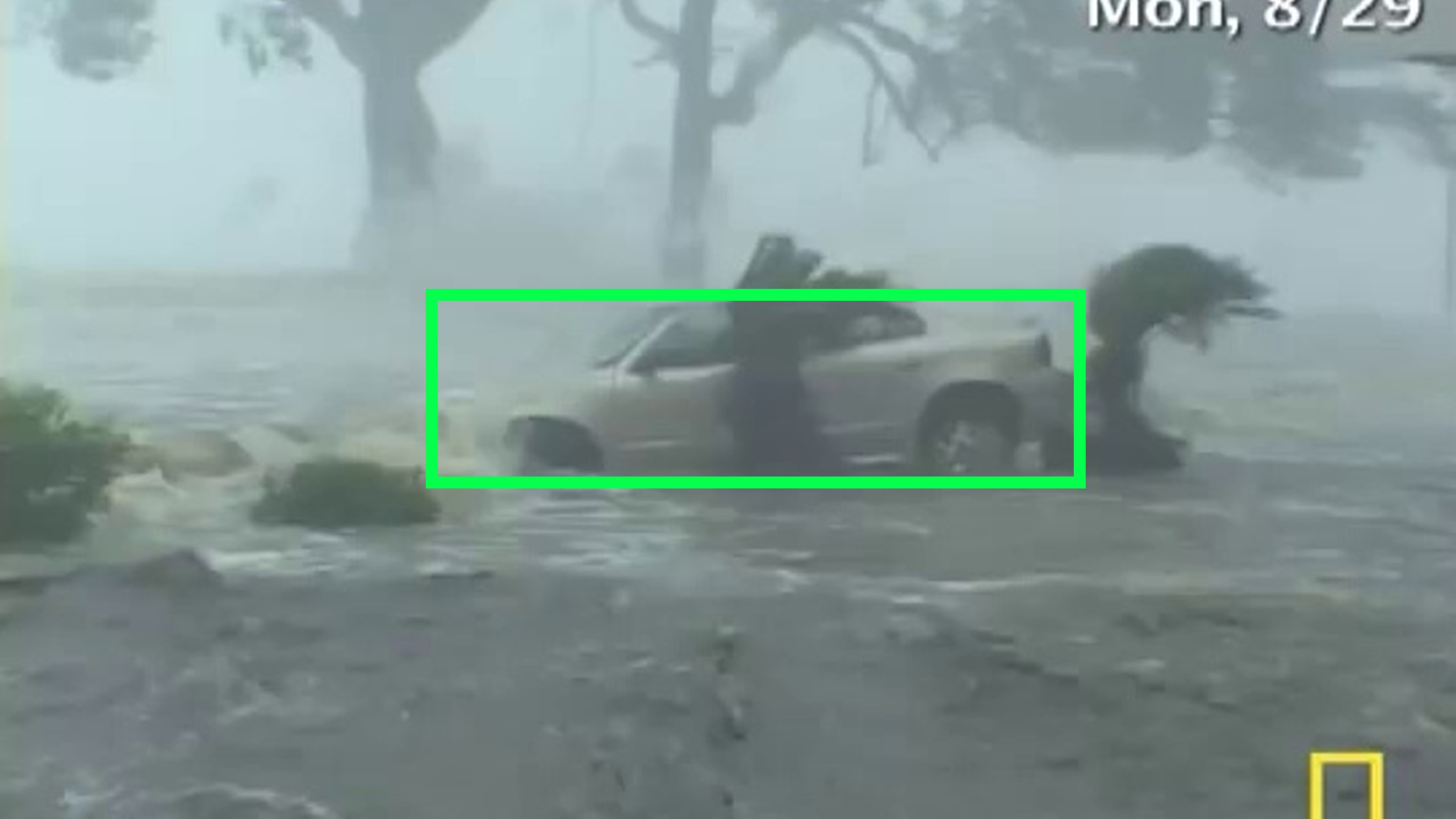}&
            \includegraphics[width=0.13\textwidth]{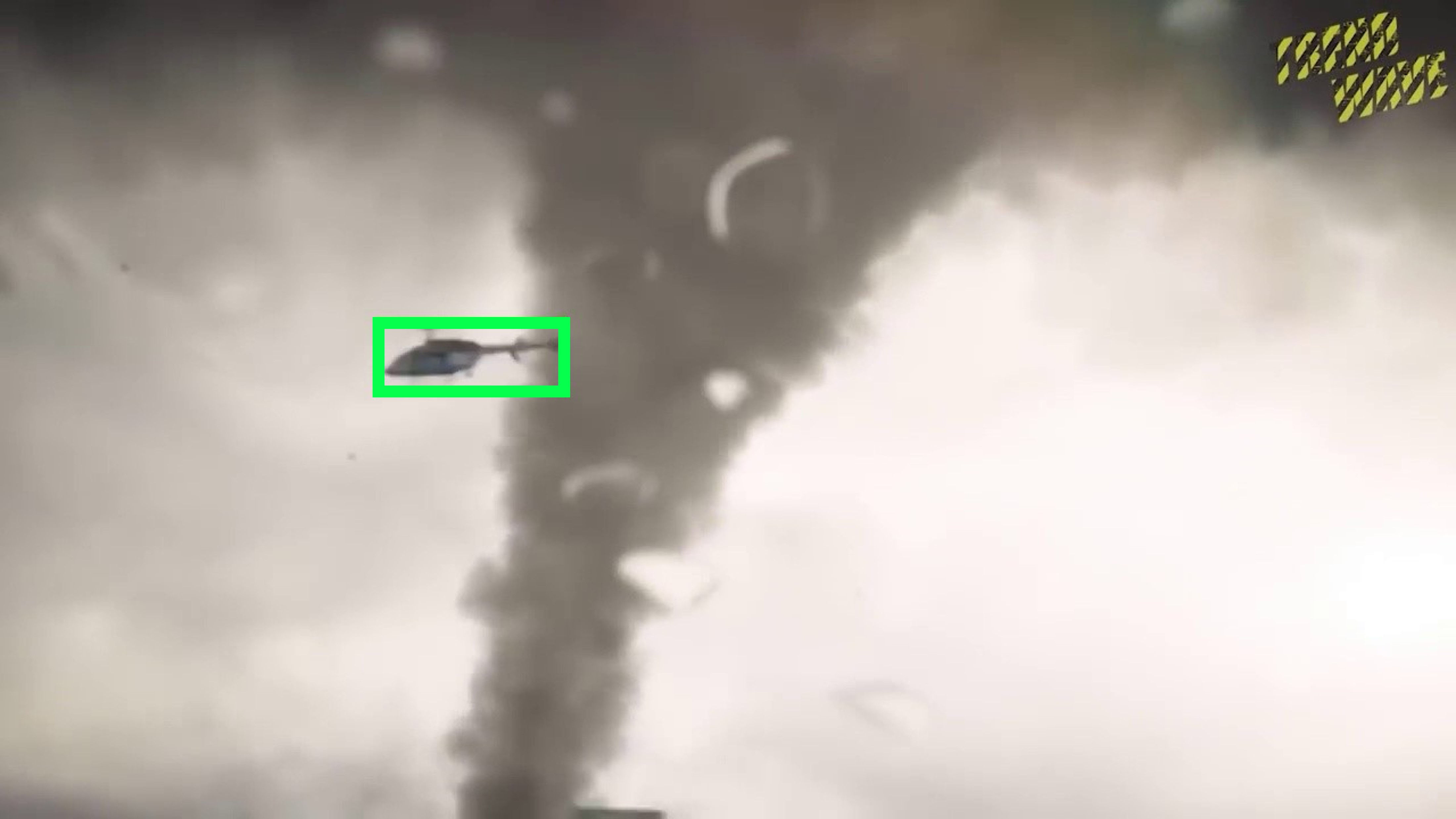} \\
            Rain & Snow & Fog & Sandstorm & Hurricane & Tornado \\
            
            \includegraphics[width=0.13\textwidth]{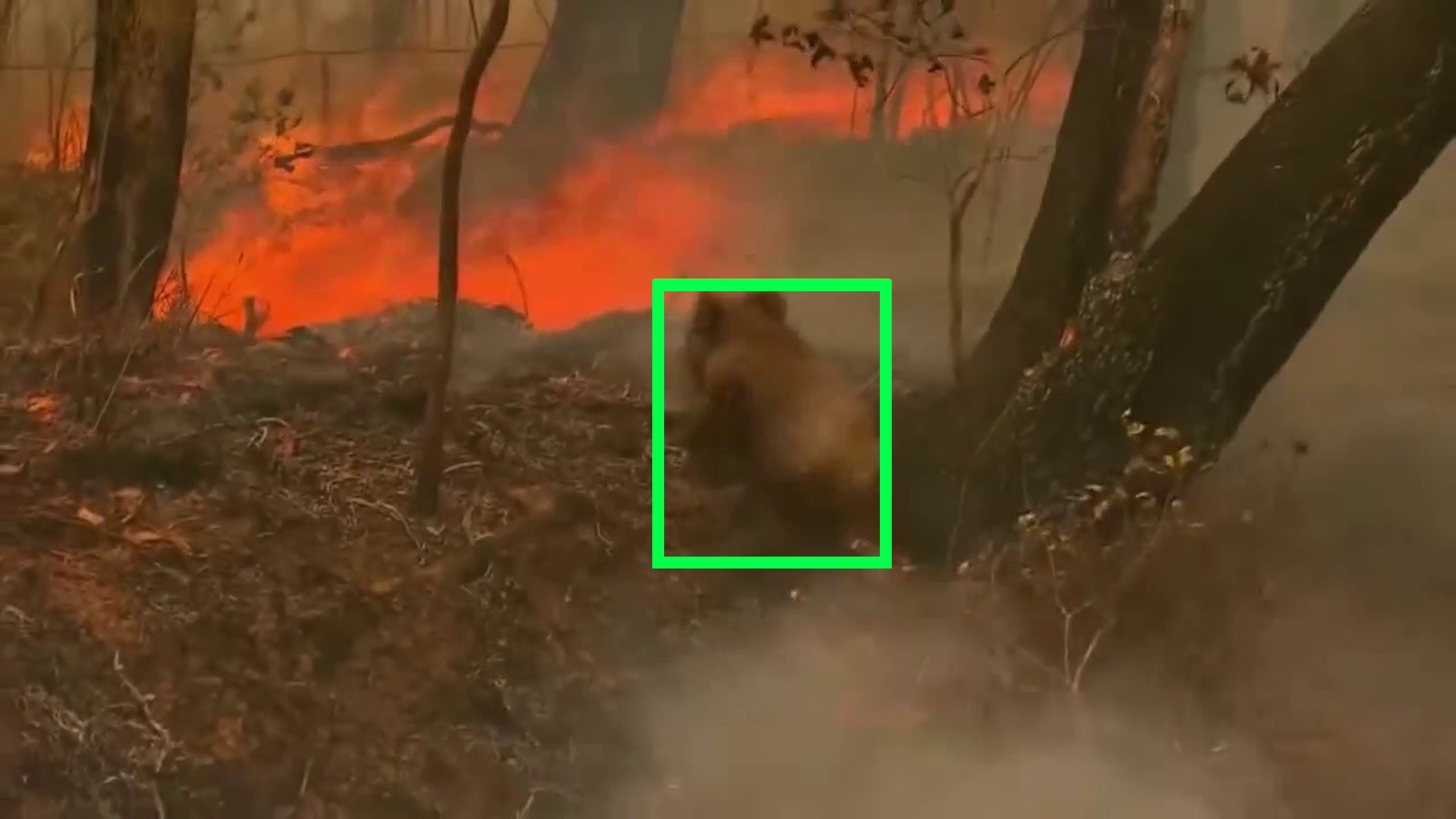}&
            \includegraphics[width=0.13\textwidth]{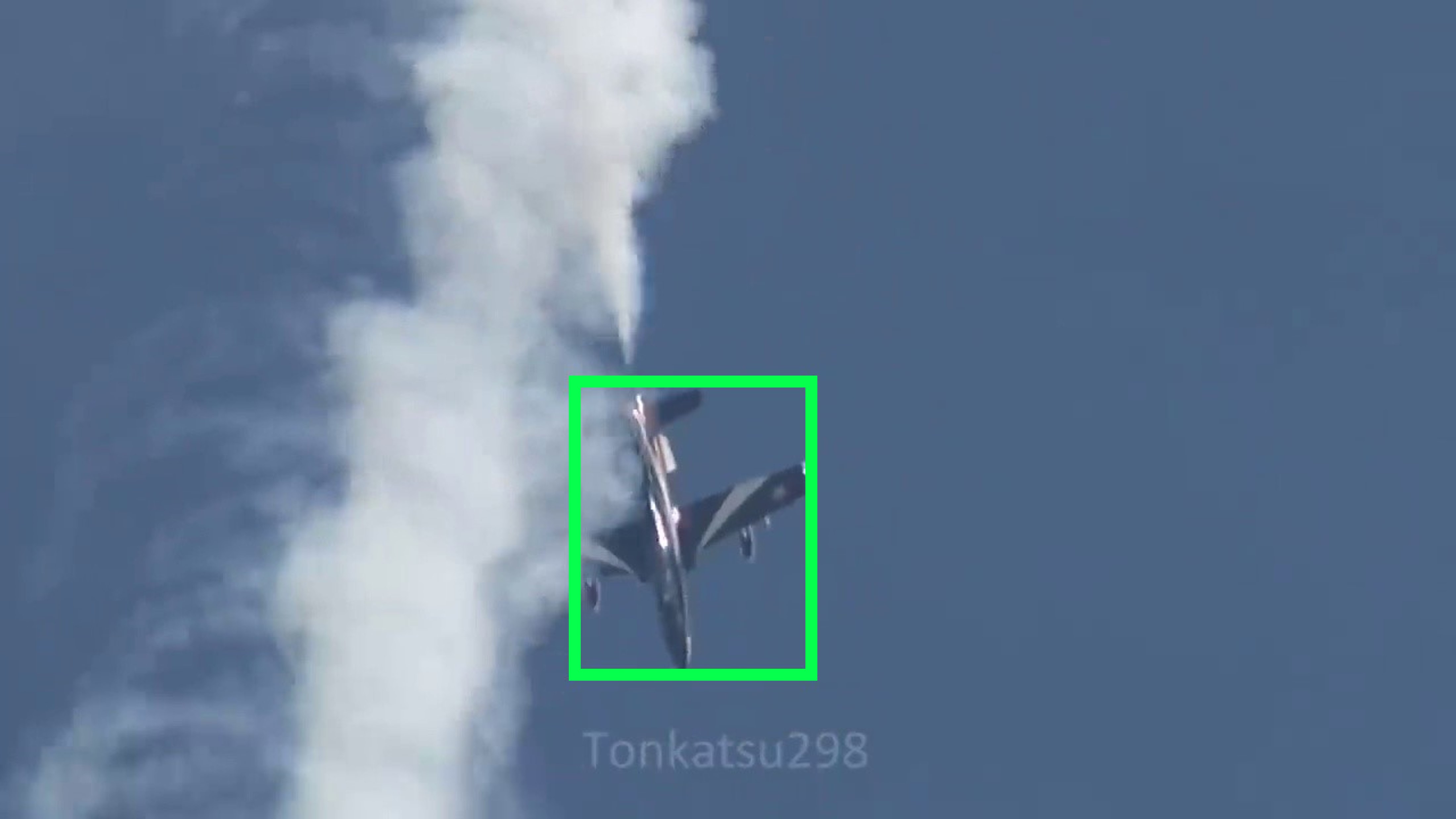}&
            \includegraphics[width=0.13\textwidth]{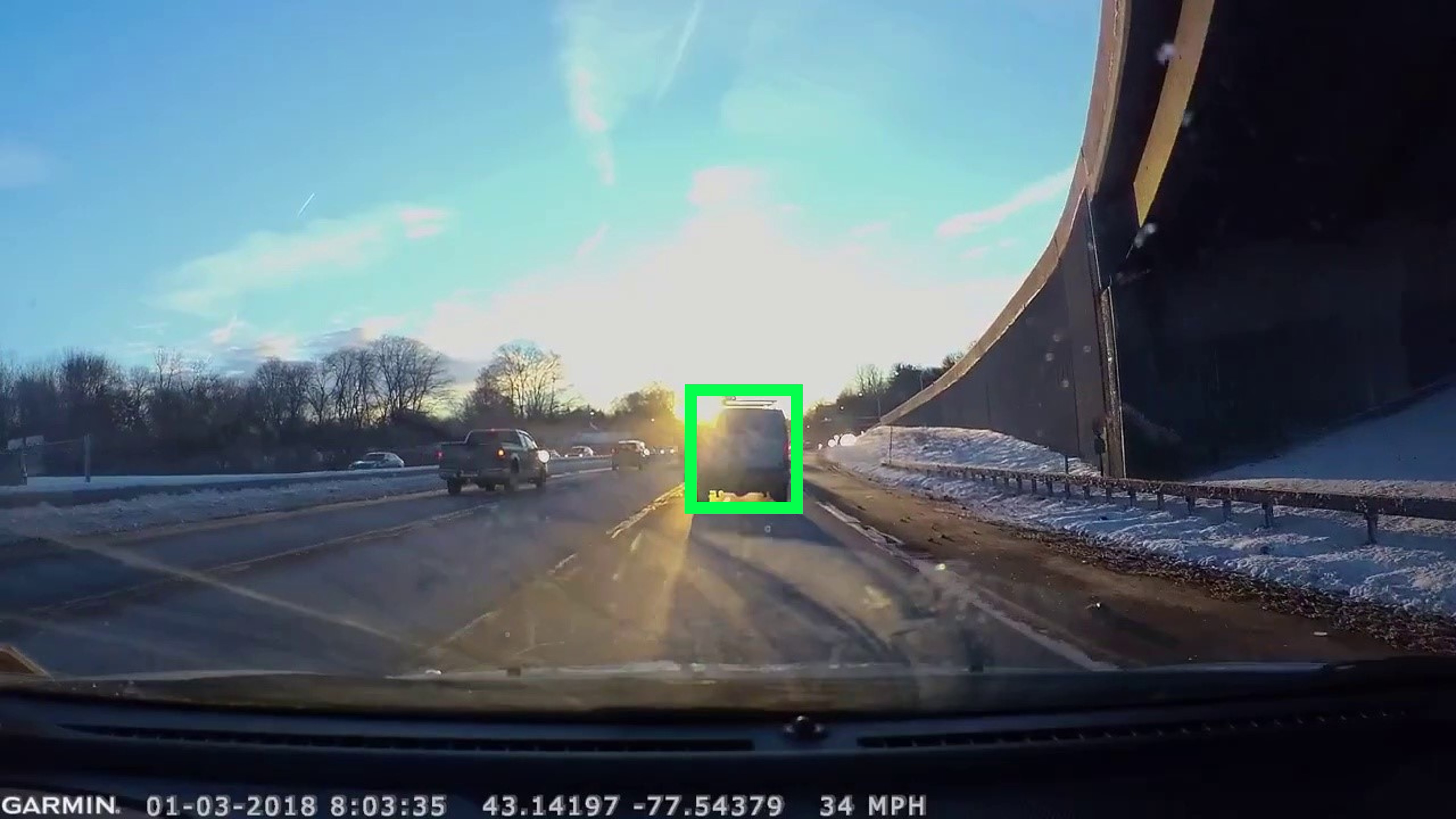}&
            \includegraphics[width=0.13\textwidth]{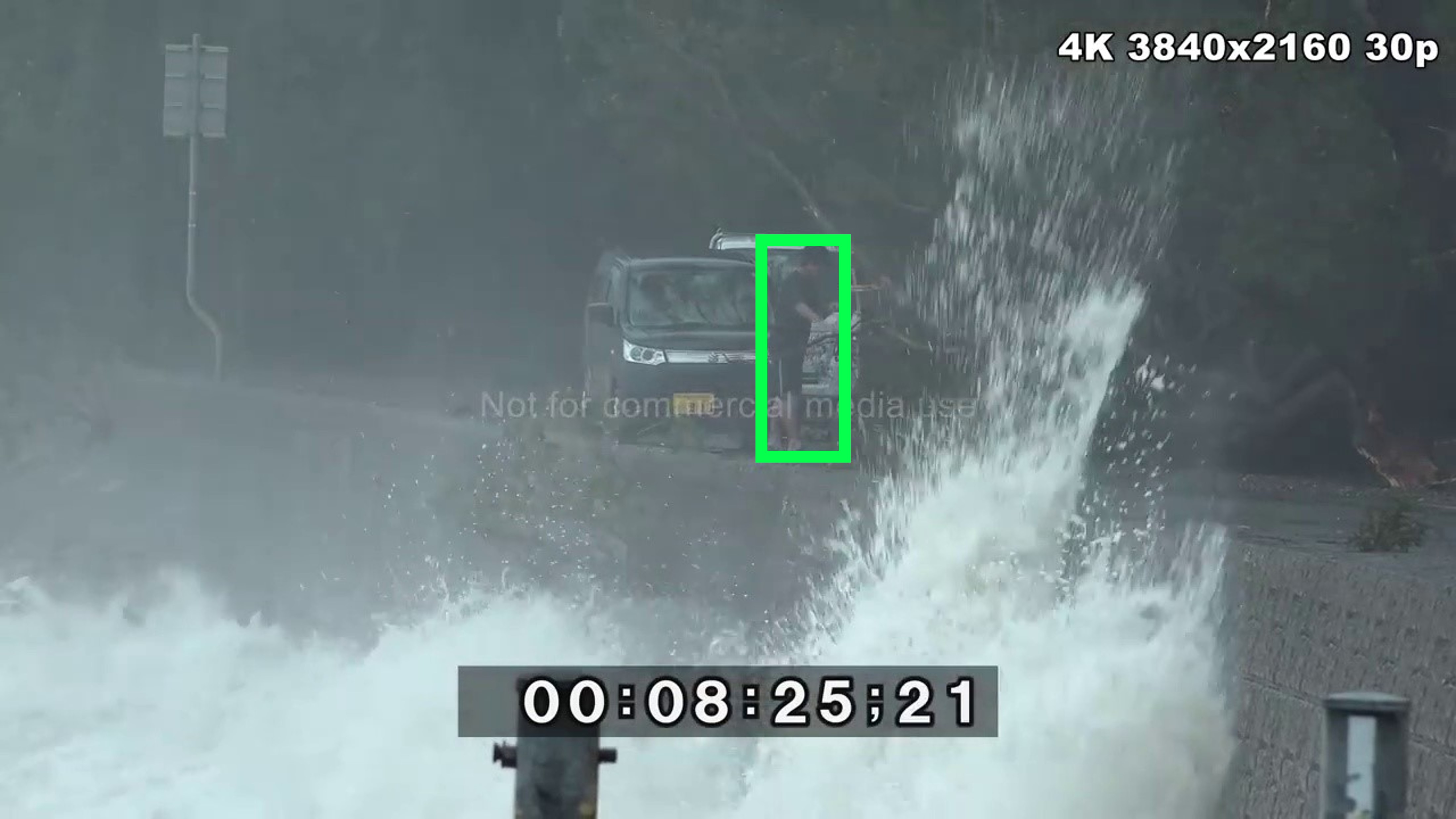}&
            \includegraphics[width=0.13\textwidth]{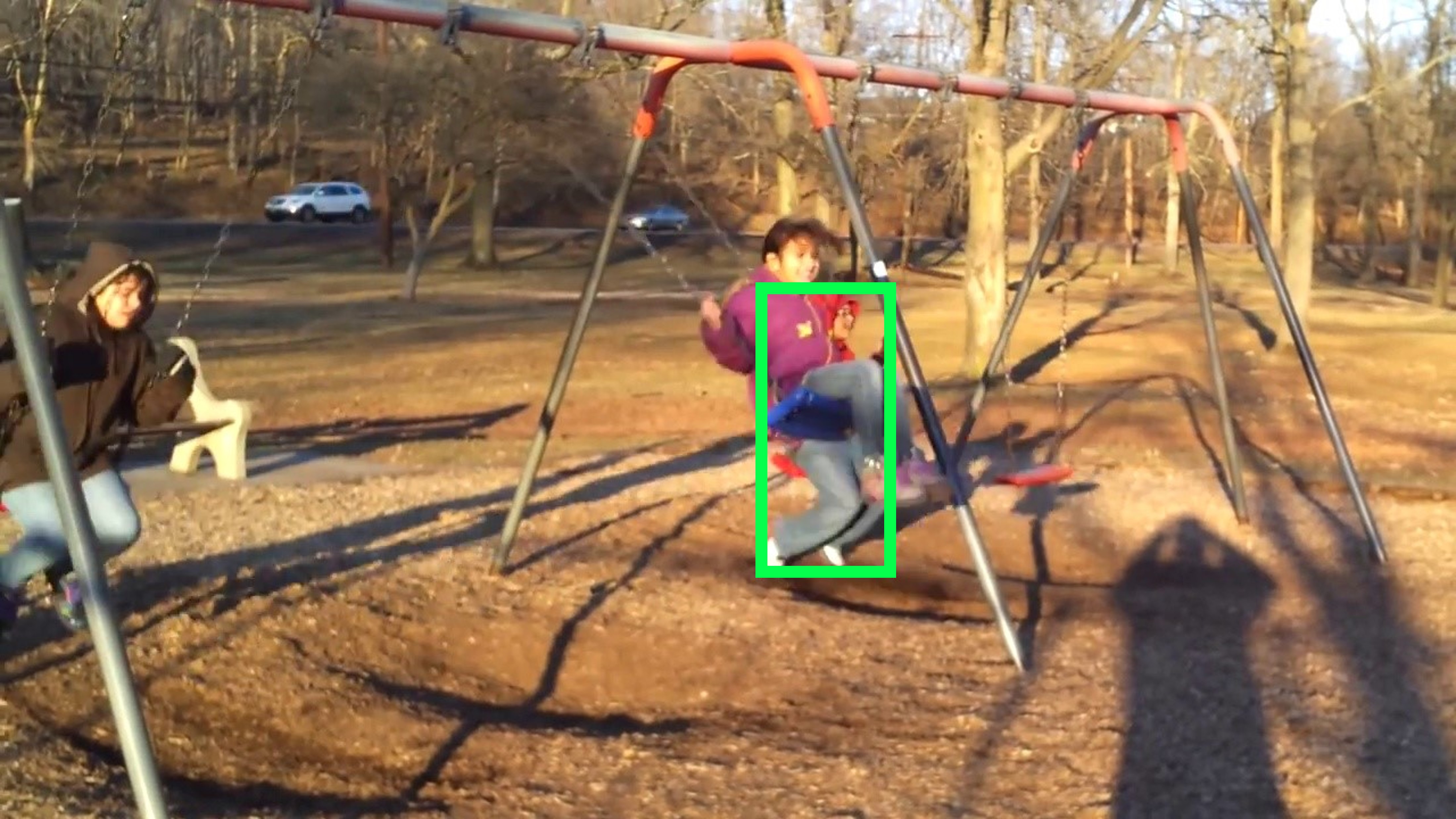}&
            \includegraphics[width=0.13\textwidth]{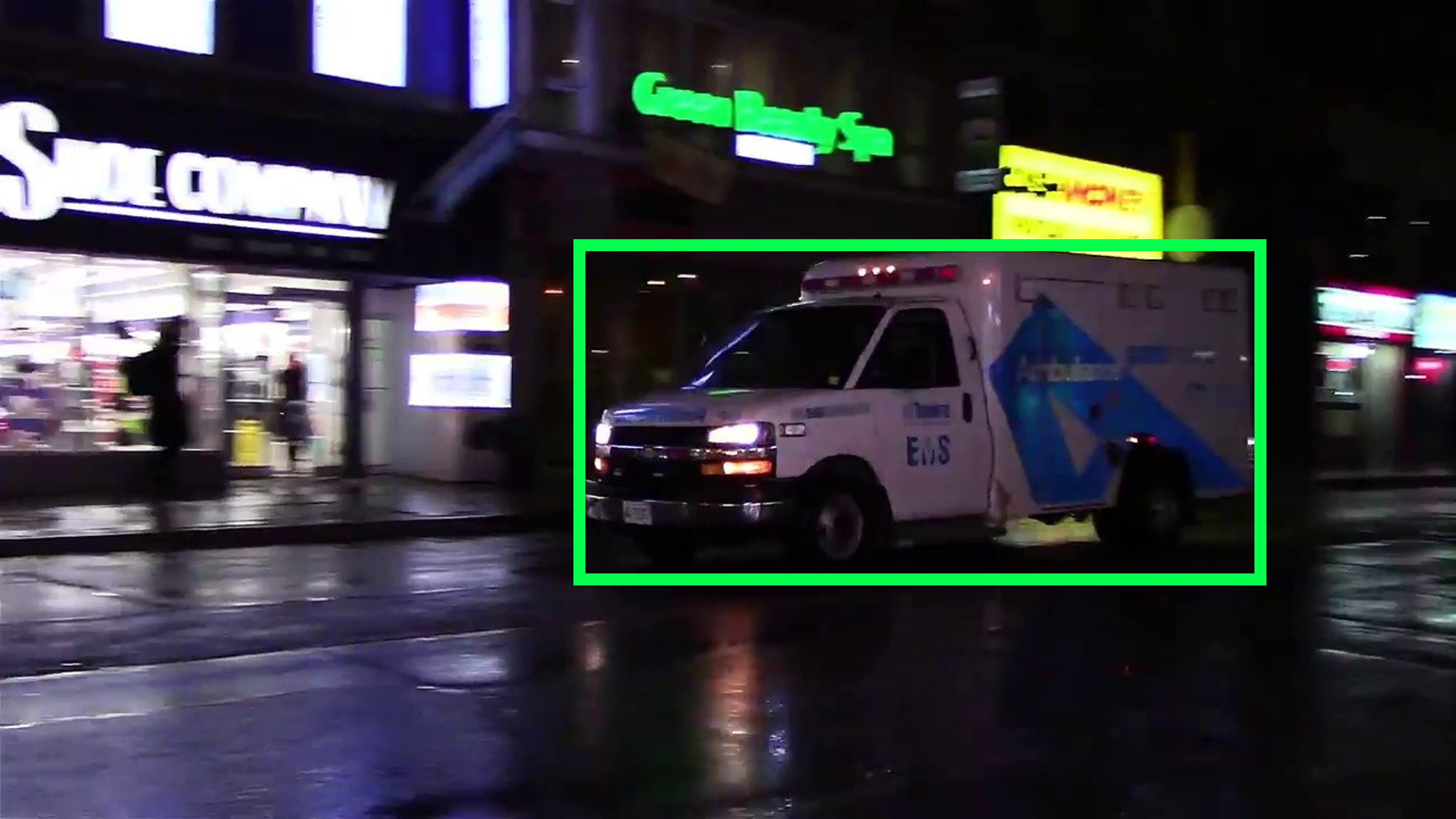} \\
            Fire & Smoke & Sun-glare & Splashing water & Occlusion & Low-light \\
            
            \includegraphics[width=0.13\textwidth]{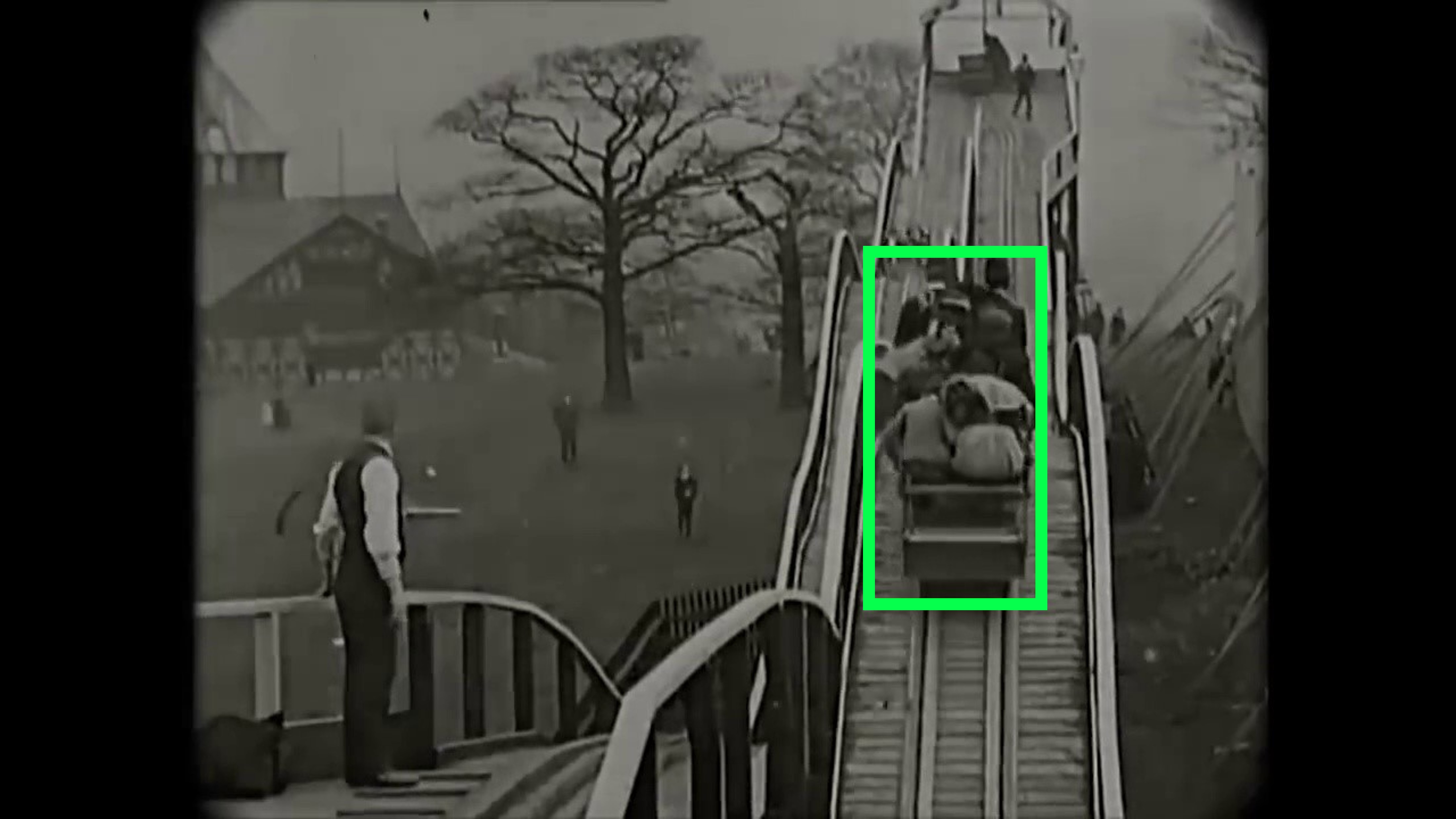}&
            \includegraphics[width=0.13\textwidth]{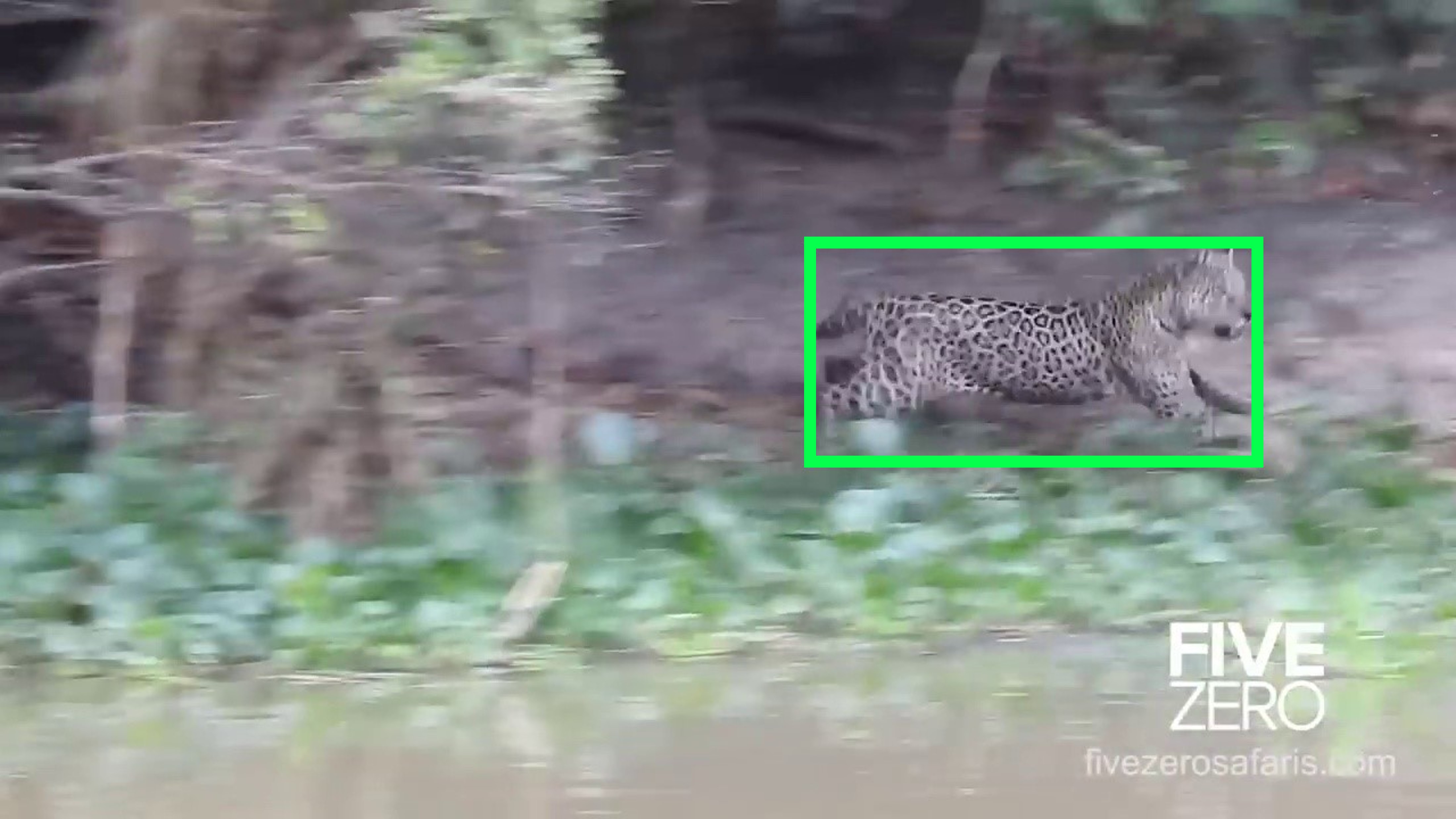}&
            \includegraphics[width=0.13\textwidth]{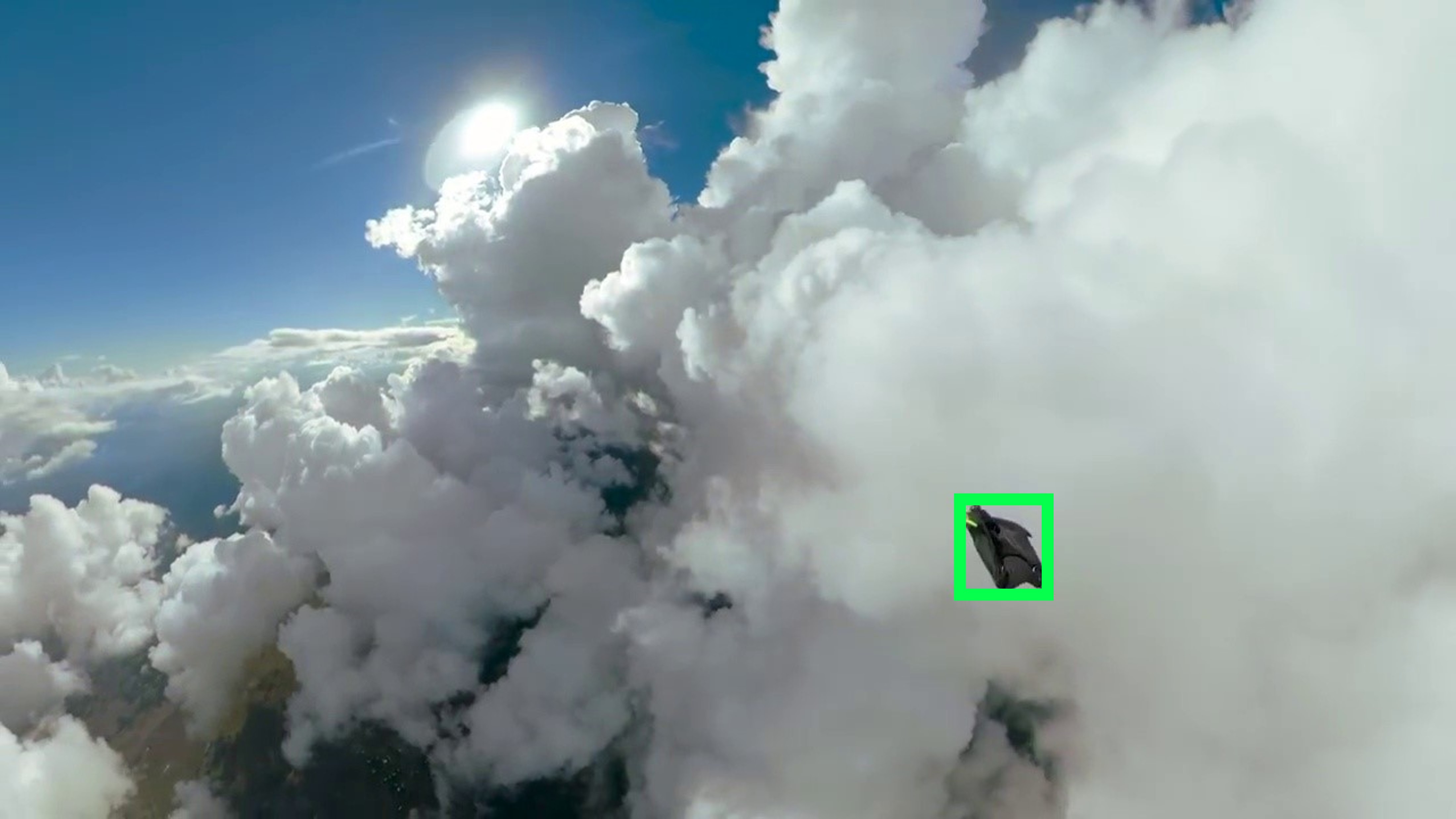}&
            \includegraphics[width=0.13\textwidth]{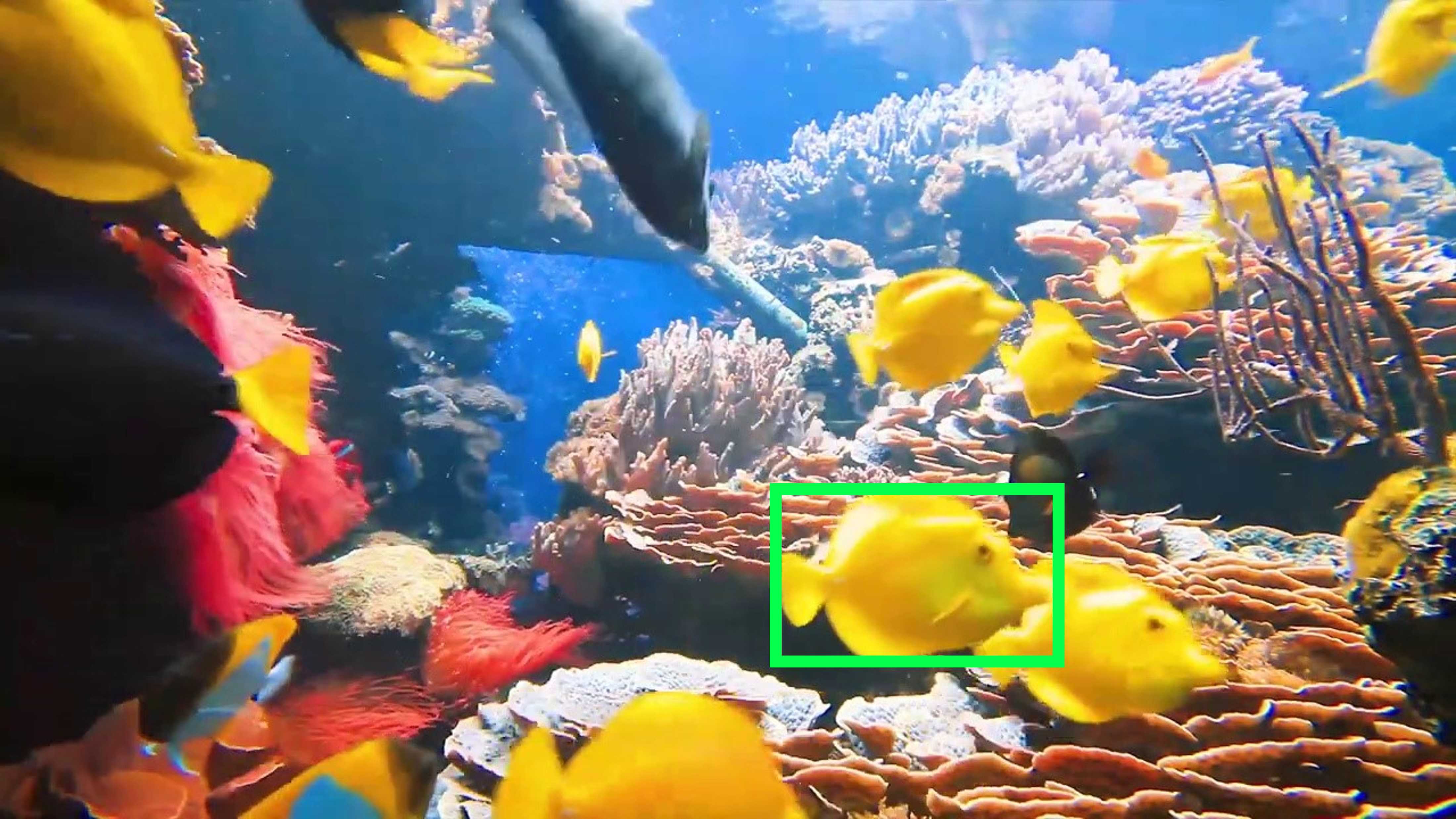}&
            \includegraphics[width=0.13\textwidth]{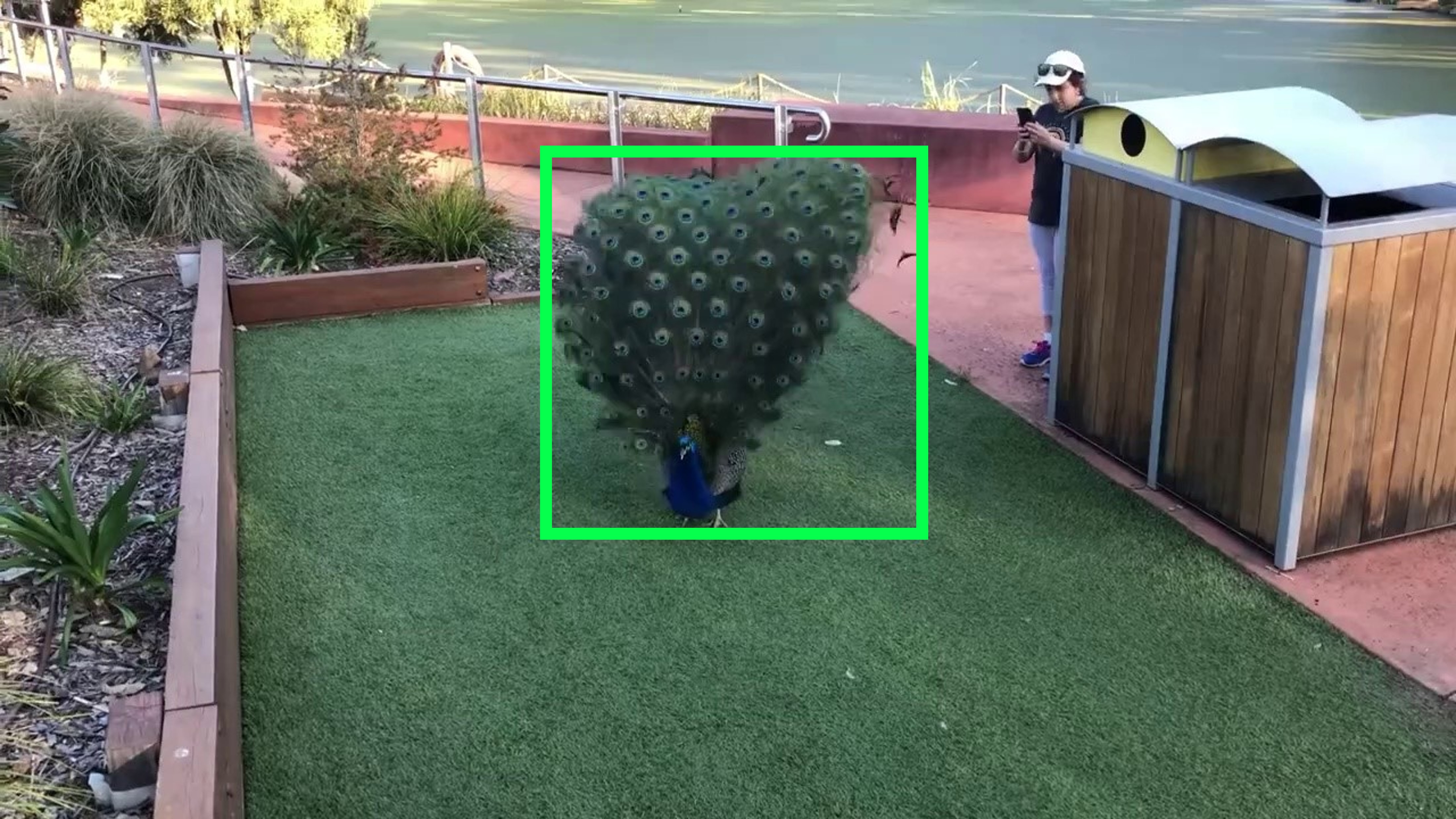}&
            \includegraphics[width=0.13\textwidth]{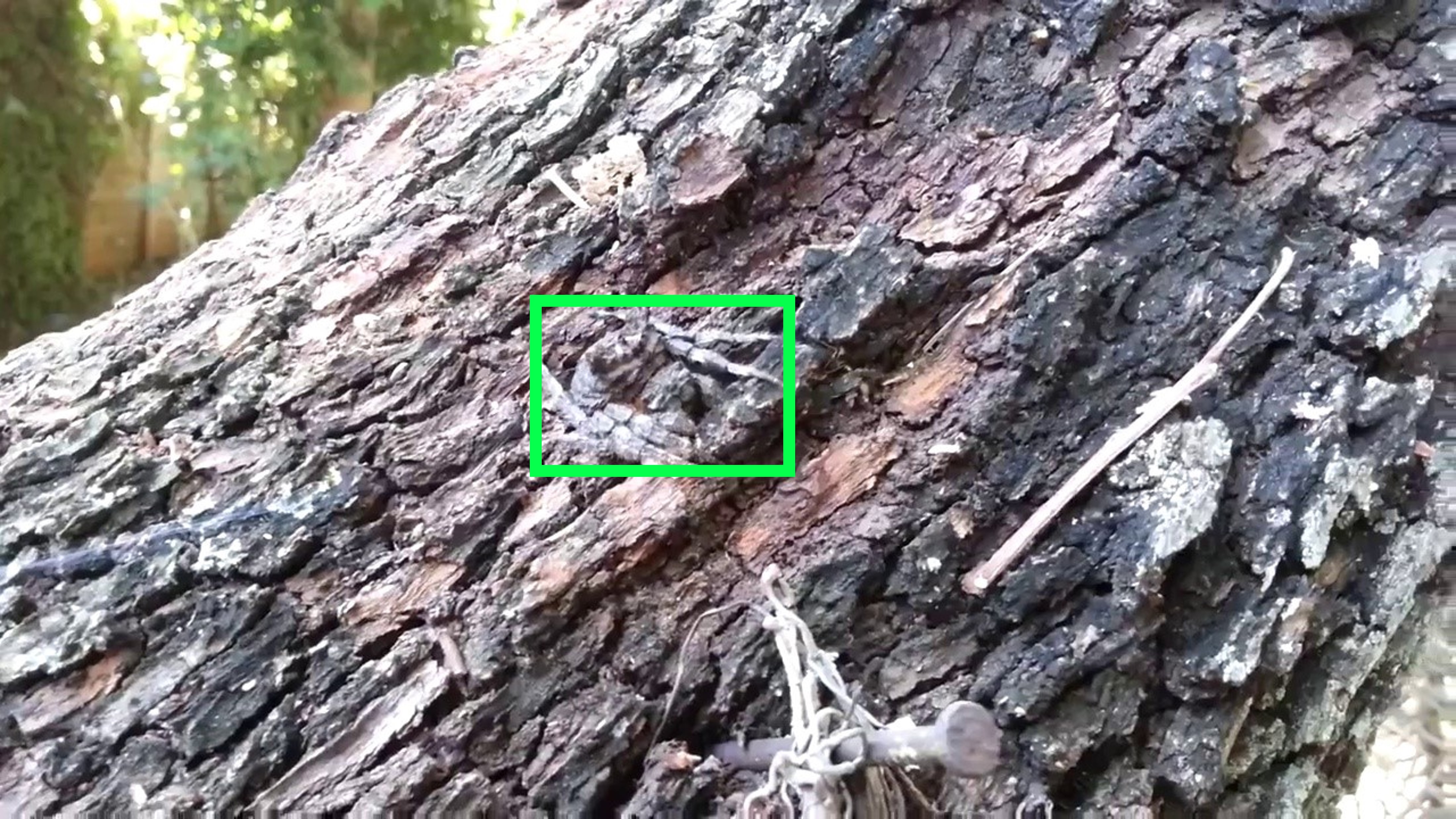} \\
            Archival & Fast motion & Small target & Distractor objects & Deformation & Camouflage \\
            & & & & &
        \end{tabular}
    \caption{Video frames corresponding to different attributes namely Rain, Snow, Fog, Sandstorm, Hurricane, Tornado, Fire, Smoke, Sun glare, Splashing water, Occlusion, Low-light, Archival video, Fast motion, Small target, Distractor objects, Deformation, Camouflage.}
    \label{fig:all_attr_exemple}
\end{figure}

\noindent\textbf{Weather Conditions:} While most existing benchmarks, such as LaSOT comprises sequences acquired under normal weather conditions, the tracking problem becomes more challenging in adverse weather scenarios, which often lead to bad visibility. In our dataset, the diverse adverse scenarios caused by weather conditions are: rain, fog, hurricane, snow, sandstorm and tornado. Accurately capturing the target appearance information in such extreme scenarios (see Fig.~\ref{fig:all_attr_exemple}) is crucial and poses additional challenges to the tracking model.

\noindent\textbf{Obstruction Effects:} Apart from difficult weather conditions, there are several  real-world  obstructions that pose additional challenges to the tracker. These obstructions can be caused by occlusion as well as natural phenomenon such as, fire, smoke, sun glare and splashing water. Our dataset covers all these diverse settings of obstructions (see Fig.~\ref{fig:all_attr_exemple}).

\noindent\textbf{Imaging Effects:} Challenging imaging conditions such as, low-light, night-time and archival monochrome videos causes losing the natural color of the target, and thereby pose difficulties to the tracking model. Our AVisT dataset comprises a set of challenging sequences covering these imaging effects (see Fig.~\ref{fig:all_attr_exemple}).

\noindent\textbf{Target Effects:} In addition to the aforementioned adverse scene scenarios, there are several target-related challenges in the real-world.  The proposed AVisT dataset includes various target effects such as, fast motion, small objects, deformations, and distractor objects (see Fig.~\ref{fig:all_attr_exemple}).

\noindent\textbf{Camouflage:} Camouflage aims to conceal the object by making it blend into the background appearance. Most animal species utilize camouflage to various degrees, with some even changing their camouflage with the seasons. This cryptic coloration makes the target hard to distinguish from the surroundings. Compared to most existing tracking benchmarks, our dataset comprises a dedicated set of  camouflage sequences that pose difficulties to state-of-the-art trackers (see Fig.~\ref{fig:all_attr_exemple}).

\subsection{Data Collection}
As discussed earlier, AVisT aims to provide a benchmark for evaluating visual trackers under diverse scenarios with adverse visibility, such as severe weather conditions, image, obstruction and target effects as well as camouflage. In addition, AVisT strives to achieve  diversity with respect to  target object classes (42 object categories in our dataset). With this objective, we first collect a large pool of around 400 videos from Youtube covering the 18 diverse scenarios with adverse visibility. We filter out  unrelated contents in each video and retain the relevant clip for tracking. We then annotate the target object in the first frame of each trimmed video. Next, we qualitatively analyze two recent representative trackers, KeepTrack~\cite{Mayer_2021_ICCV_KeepTrack} and STARK~\cite{Yan_2021_ICCV_STARK},  on these image sequences. We select a set of 120 sequences which are highly challenging for both these recent representative trackers. We note that other trackers such as, ToMP~\cite{Mayer_2022_CVPR_Tomp}, and MixFormer-1k \cite{Cui_2022_CVPR_Mixformer},  which perform similar to KeepTrack and STARK on LaSOT~\cite{Fan_2019_CVPR_Lasot}, also perform similarly on our AVisT dataset. Therefore, the 120 videos we select are  generally challenging and not specific to these two representative trackers. 
Among the initially large pool of around 400 videos, some of the camouflage sequences are overlapping with the MoCA dataset~\cite{MoCA}. However, we re-annotate those camouflage sequences since the MoCA dataset was originally proposed for camouflage object detection and hence does not provide dense frame-level annotations required for visual object tracking.  

To summarize, our AVisT dataset comprises 120  challenging videos from YouTube under the Creative Commons licence. All these 120  videos belong to at least one of the  diverse scenarios, with a total of 80k annotated video frames. The frame-rates of these videos ranges from 24 to 30 frames per second~(fps) and the average sequence length is 664 frames \textcolor {black} {(i.e., 22.2 seconds with 30~fps)}.  The shortest sequence in our dataset has 99 frames (3.3 seconds with 30~fps), while the longest one has 3113 frames (103.7 seconds with 30~fps). 

\subsection{Dataset Annotation}
After data collection, the next step is to obtain high-quality annotations for all the sequences to ensure an accurate benchmarking of visual trackers.
To obtain consistent annotations, we standardize a protocol that ensures high-quality annotations for the proposed AVisT dataset. During the annotation process, a video is processed by two teams, where the labeling team, comprising typically three members, manually draws the target object's bounding box as the tightest axis-aligned rectangle that fits the target in each frame of a video that has a specified tracking target. Afterwards, the validation team reviews the annotation results with either unanimously  agreeing on the annotation results or returning it back to the labeling team for revising the annotation.

\begin{figure}[t]
  \centering
    \includegraphics[width=0.65\linewidth]{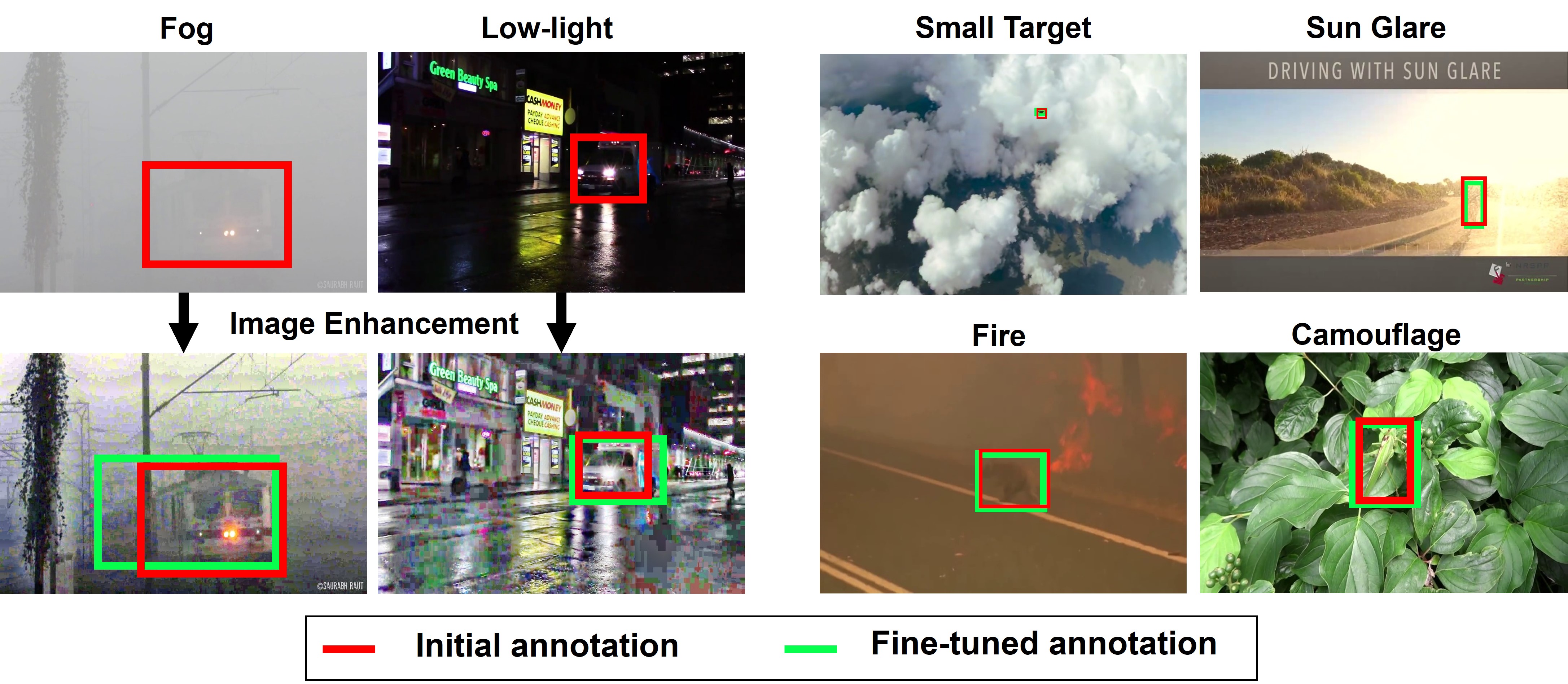}
  \caption{Example frames from diverse scenarios highlighting the complexity of the annotation process. In case of severe visibility, we employ image enhancement techniques to better distinguish the target aiding in improved annotations. Here, the example frames of different scenarios show that fine-tuned annotations (green color) better fit the target object region, compared to the initial annotations (red color).}\label{fig:annotations}
\end{figure}
\noindent\textbf{Quality Control:} In case of diverse scenarios where the target object suffers from extreme visibility issues (\eg, dense fog and low-light), the annotation process becomes further challenging. Therefore, we employ standard image enhancement techniques, including contrast limited adaptive histogram equalization, histogram equalization, gamma correction, brightness and contrast adjustment, and white balance, to distinguish the boundary of the target and improve the annotation quality (see Fig.~\ref{fig:annotations} for fog and low-light examples). Furthermore, we also utilize the relative displacement of the target between consecutive frames to estimate its boundaries. As discussed earlier, we also improve the annotation quality by making separate teams for the process of labeling and validation. The labeling team manually annotates and cross checks the annotated video samples. Afterwards, the validation team verifies the quality of the annotated videos and points out any possible mistakes in the annotations which are then corrected by the respective team members.  
The annotation teams meticulously examine the annotations and often revise them in order to enhance the quality of the annotation. In the initial phase of validation, around 24$\%$ of the original annotations were corrected. Additionally, several frames underwent more than three revisions. Fig. \ref{fig:annotations} presents example frames where the initial annotations are fine-tuned, leading to improved annotation quality. 

Our AVisT benchmark comprises different flags, where a frame can be labeled with full occlusion, partial occlusion, out-of-view and extreme visibility. A full occlusion flag is set when the target object is fully occluded by another object, such that the original pixel values are hard to be recovered. The partial occlusion flag implies that the target object is partially visible. In such a case, we annotate only the visible part of the target in the frame. The out-of-view flag refers to the case where the target object is not in the camera field-of-view. Here, we set a dedicated flag indicating that the target object is out-of-view in the frame. The extreme visibility flag refers to the cases of dense fog and extreme low-light where the target is suffering from severe visibility issues and is hard to identify for human eyes. We observe that image enhancement techniques helps in improving the annotation process in such cases. %

\section{Experiments}
\subsection{Evaluated Trackers}

The field of generic tracking has greatly progressed in recent years with the development of various approaches. Siamese trackers employ a deep network to extract a target template that is matched with the features of the current video frame in order to localize the target therein. In contrast, trackers based on a discriminative classifier learn the weights of a convolutional kernel that allows to differentiate between the target object (foreground) and background regions in the current video frame. More recently, transformer-based trackers have emerged that use self and cross attention layers to combine template and search frame information to extract discriminative features to localize the target. 
To analyse our AVisT, we evaluate high-performance and popular trackers that we briefly summarize bellow.

\begin{table}[t]
	\newcommand{\best}[1]{\textbf{\textcolor{red}{#1}}}
	\newcommand{\scnd}[1]{\textbf{\textcolor{blue}{#1}}}
	\newcommand{\opt}[1]{\textbf{\textcolor{violet}{#1}}}
	\newcommand{\fast}[1]{\textbf{\textcolor{orange}{#1}}}
	\newcommand{\yes}{\textcolor{black}{\checkmark}}
	\newcommand{\no}{\textcolor{black}{\ding{55}}}
	\newcommand{\dist}{\hspace{12pt}}%
	\begin{center}
    	\resizebox{0.86\linewidth}{!}{%
            \begin{tabular}{l@{\dist}l@{\dist}c@{\dist}c@{\dist}c@{\dist}c@{\dist}c@{\dist}c@{\dist}}
            	\toprule
            	Framework                & Name                                                 & Backbone   & Online Update & Venue & Success (AUC) & OP50 & OP75 \\
            	\midrule
                \multirow{4}{*}{Siamese} & SiamMask~\cite{Wang_2019_CVPR_SiamMask}              & ResNet-50  & \no  & CVPR 2019 & 35.75 & 40.06 & 18.45 \\
                                         & SiamRPN++~\cite{Li_2019_CVPR_SiamRPN++}              & ResNet-50  & \no  & CVPR 2019 & 39.01 & 43.48 & 21.18 \\                                
                                         & SiamBAN~\cite{Chen_2020_CVPR_SiamBAN}                & ResNet-50  & \no  & CVPR 2020 & 37.58 & 43.22 & 21.73 \\
                                         & Ocean~\cite{Zhang_2020_ECCV_Ocean}                   & ResNet-50  & \yes & ECCV 2020 & 38.89 & 43.60 & 20.47 \\
                \midrule
                \multirow{9}{*}{\shortstack{Discriminative\\Classifier}} & Atom~\cite{Danelljan_2019_CVPR_ATOM}               & ResNet-18 & \yes & CVPR 2019 & 38.61 & 41.51 & 22.17 \\
                                                                         & DiMP-18~\cite{Bhat_2019_ICCV_DIMP}                 & ResNet-18 & \yes & ICCV 2019 & 40.55 & 44.07 & 23.67 \\
                                                                         & DiMP-50~\cite{Bhat_2019_ICCV_DIMP}                 & ResNet-50 & \yes & ICCV 2019 & 41.91 & 45.67 & 25.95 \\                                 
                                                                         & PrDiMP-18~\cite{Danelljan_2020_CVPR_PRDIMP}        & ResNet-18 & \yes & CVPR 2020 & 41.65 & 45.80 & 27.20 \\
                                                                         & PrDiMP-50~\cite{Danelljan_2020_CVPR_PRDIMP}        & ResNet-50 & \yes & CVPR 2020 & 43.25 & 48.02 & 28.70 \\
                                                                         & Super DiMP~\cite{Danelljan_2019_github_pytracking} & ResNet-50 & \yes & CVPR 2020 & 48.39 & 54.61 & 33.99 \\
                                                                         & KYS~\cite{Bhat_2020_ECCV_KYS}                      & ResNet-50 & \yes & ECCV 2020 & 42.53 & 46.67 & 26.83 \\
                                                                         & KeepTrack~\cite{Mayer_2021_ICCV_KeepTrack}         & ResNet-50 & \yes & ICCV 2021 & 49.44 & 56.25 & 37.75 \\
                                                                         & AlphaRefine~\cite{Yan_2021_CVPR_AlphaRefine}       & ResNet-50 & \yes & CVPR 2021 & 49.63 & 55.65 & 38.17 \\
                \midrule
                \multirow{10}{*}{Transformer}   & TrSiam~\cite{Wang_2021_CVPR_TrDiMP}            & ResNet-50  & \yes & CVPR 2021 & 47.82 & 54.84 & 33.04 \\
                                               & TrDiMP~\cite{Wang_2021_CVPR_TrDiMP}             & ResNet-50  & \yes & CVPR 2021 & 48.14 & 55.26 & 33.77 \\
                                               & TransT~\cite{Chen_2021_CVPR_TransT}             & ResNet-50   & \no  & CVPR 2021 & 49.03 & 56.43 & 37.19 \\
                                               & STARK-ST-50~\cite{Yan_2021_ICCV_STARK}          & ResNet-50  & \yes & ICCV 2021 & 51.11 & 59.20 & 39.07 \\
                                               & STARK-ST-101~\cite{Yan_2021_ICCV_STARK}         & ResNet-101 & \yes & ICCV 2021 & 50.50 & 58.23 & 38.97 \\
                                               & ToMP-50~\cite{Mayer_2022_CVPR_Tomp}             & ResNet-50  & \yes & CVPR 2022 & 51.60 & 59.47 & 38.87 \\
                                               & ToMP-101~\cite{Mayer_2022_CVPR_Tomp}            & ResNet-101 & \yes & CVPR 2022 & 50.90 & 58.77 & 38.42 \\
                                               & MixFormer-1k~\cite{Cui_2022_CVPR_Mixformer}     & MAM        & \yes & CVPR 2022 & 50.83 & 58.56    & 39.30  \\                                
                                               & MixFormer-22k~\cite{Cui_2022_CVPR_Mixformer}    & MAM        & \yes & CVPR 2022 & 53.72 & 62.98 & 43.02 \\ 
                                               & MixFormerL-22k~\cite{Cui_2022_CVPR_Mixformer}   & MAM        & \yes & CVPR 2022 & \textbf{55.99} & \textbf{65.92} & \textbf{46.34} \\
                \bottomrule
            \end{tabular}
    	}
	\end{center}
	\caption{Comparison of different trackers in terms of AUC score on AVisT. Other than MixFormerL-22k and MixFormer-22k, the evaluated trackers utilize backbones trained on ImageNet-1K dataset. Among existing trackers, STARK-ST-50 and ToMP-50 achieve comparable AUC scores. Both MixFormerL-22k and MixFormer-22k utilizing backbone with ImageNet-22K pre-training obtain improved tracking performance. In addition to AUC score, we also report performance at OP50 and OP75.
	}
	\label{tab:trackers}%
\end{table}

\begin{figure}[h]
\begin{center}
\begin{subfigure}{0.28\linewidth}
    \includegraphics[width=1\linewidth, keepaspectratio]{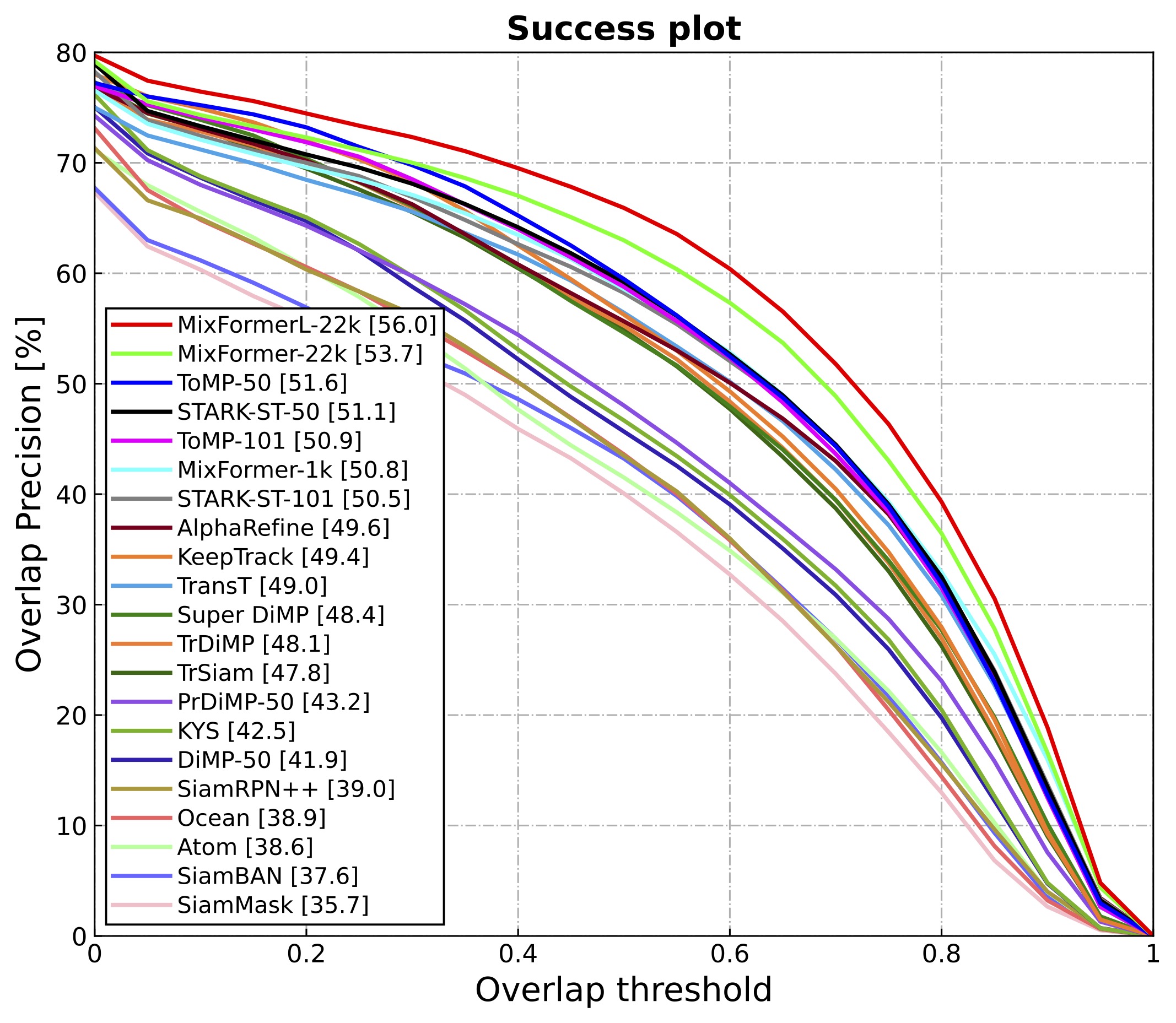}
    \vspace{-6mm}
    \caption{ Overall Results}
    \label{fig:success_wc}
  \end{subfigure}%
\begin{subfigure}{0.28\linewidth}
    \includegraphics[width=1\linewidth, keepaspectratio]{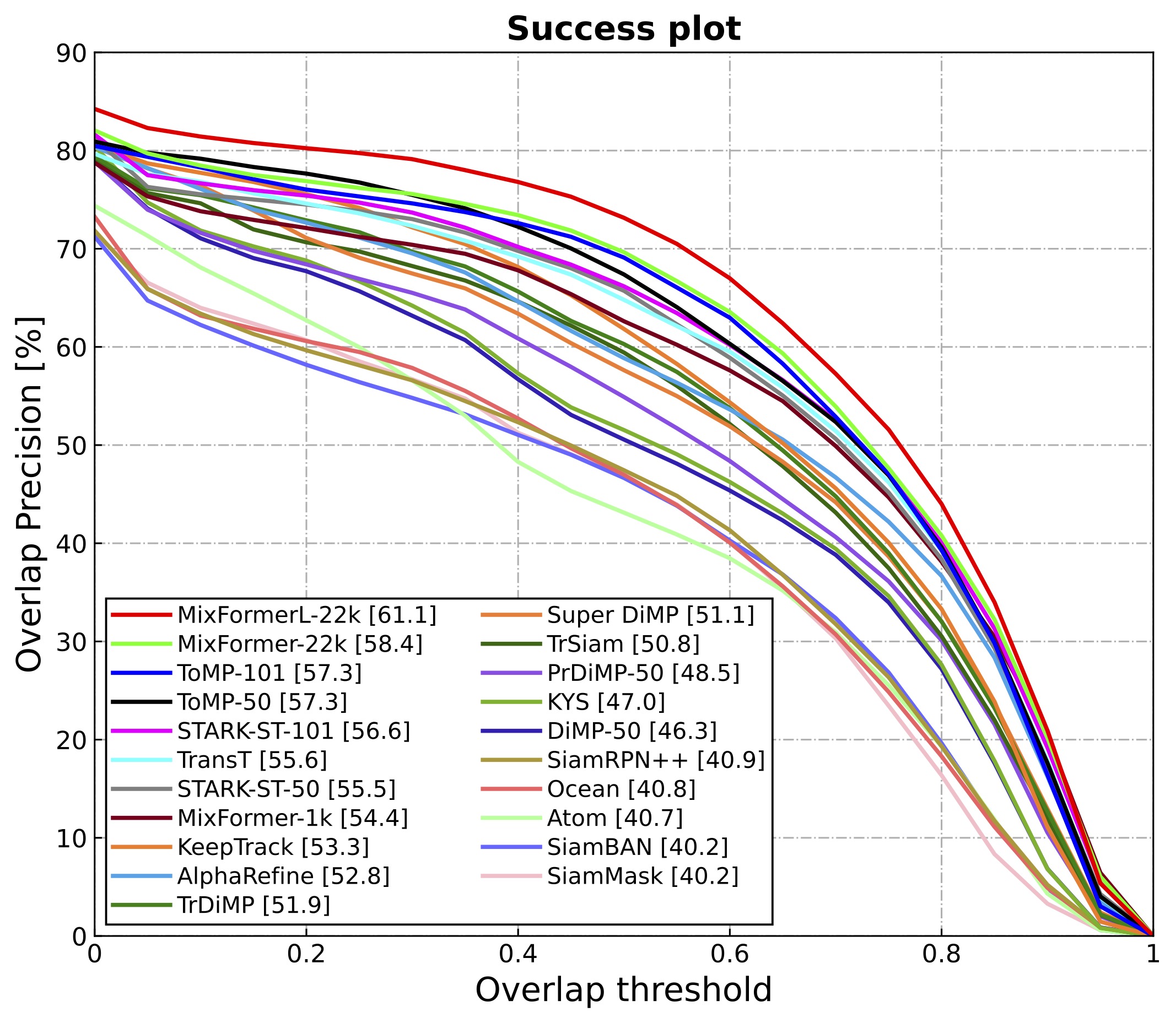}
    \vspace{-6mm}
    \caption{Weather Conditions}
    \label{fig:success_wc}
  \end{subfigure}%
  \begin{subfigure}{0.28\linewidth}
    \includegraphics[width=1\linewidth, keepaspectratio]{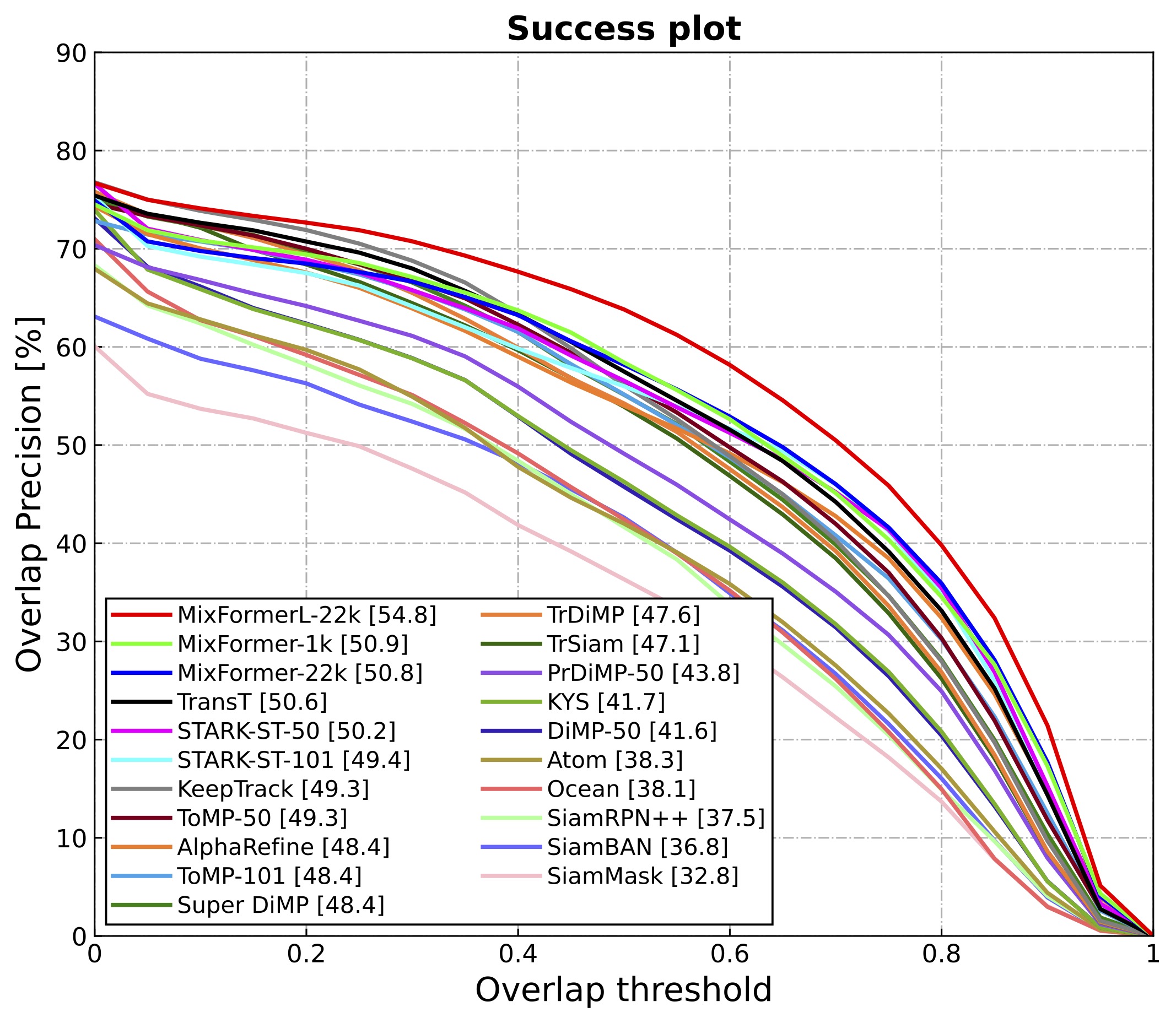}
    \vspace{-6mm}
    \caption{Obstruction Effects}
    \label{fig:success_oe}
  \end{subfigure}%
  \\
  \begin{subfigure}{0.28\linewidth}
    \includegraphics[width=1\linewidth, keepaspectratio]{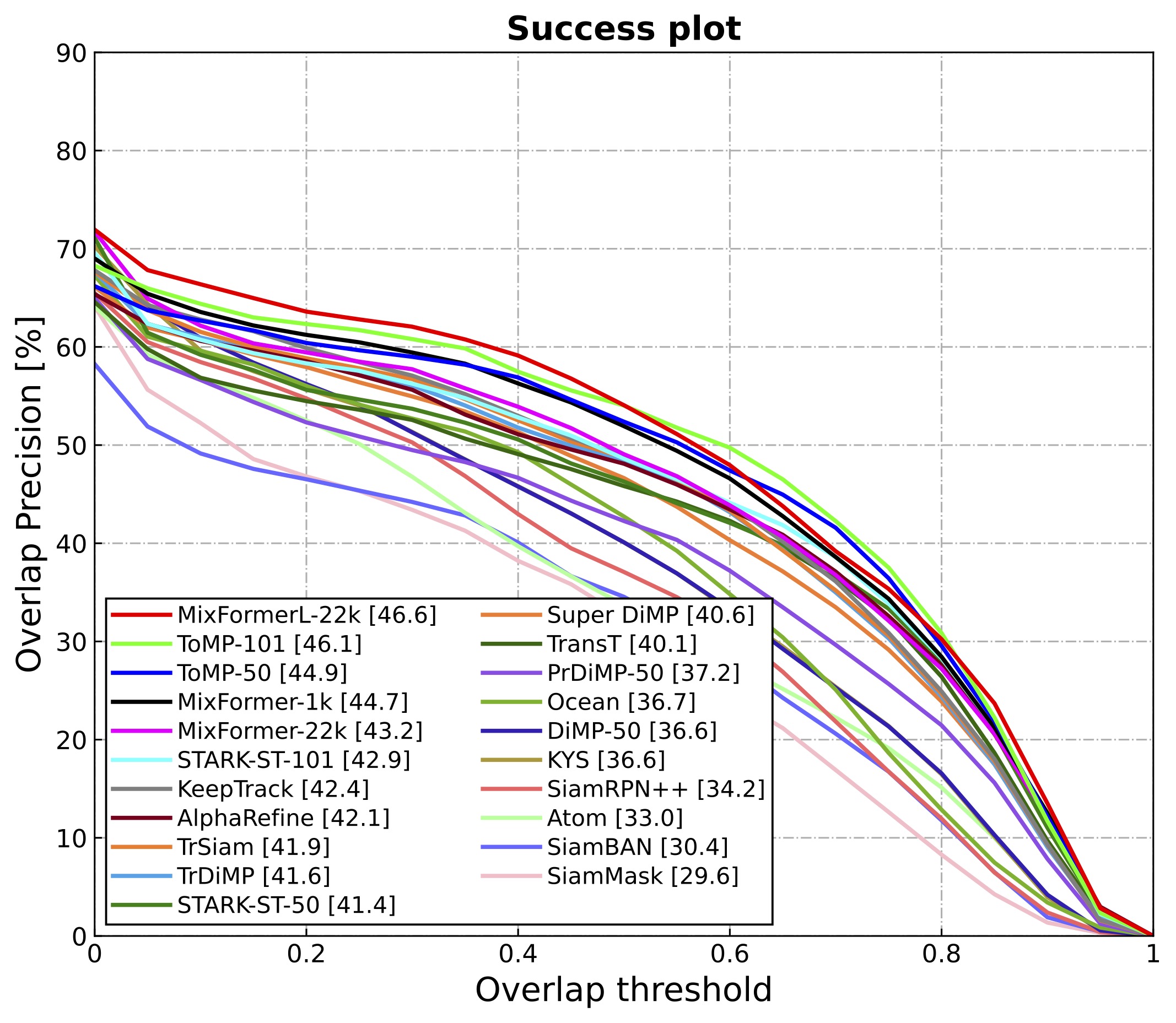}
    \vspace{-6mm}
    \caption{Imaging  Effects}
    \label{fig:success_ie}
  \end{subfigure}%
  \begin{subfigure}{0.28\linewidth}
    \includegraphics[width=\linewidth, keepaspectratio]{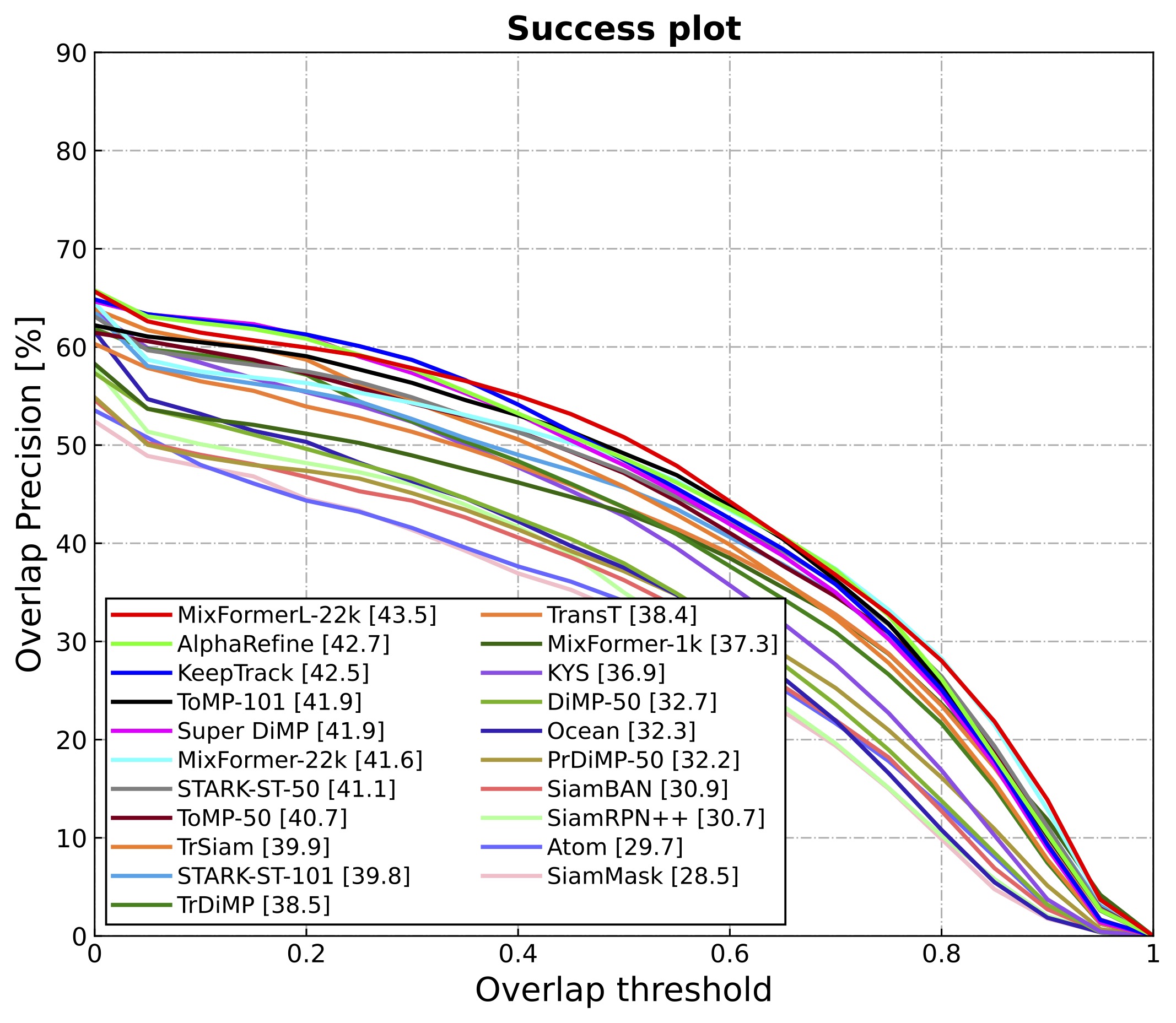}
    \vspace{-6mm}
    \caption{Target Effects}
    \label{fig:success_te}
  \end{subfigure}%
  \begin{subfigure}{0.28\linewidth}
    \includegraphics[width=1\linewidth, keepaspectratio]{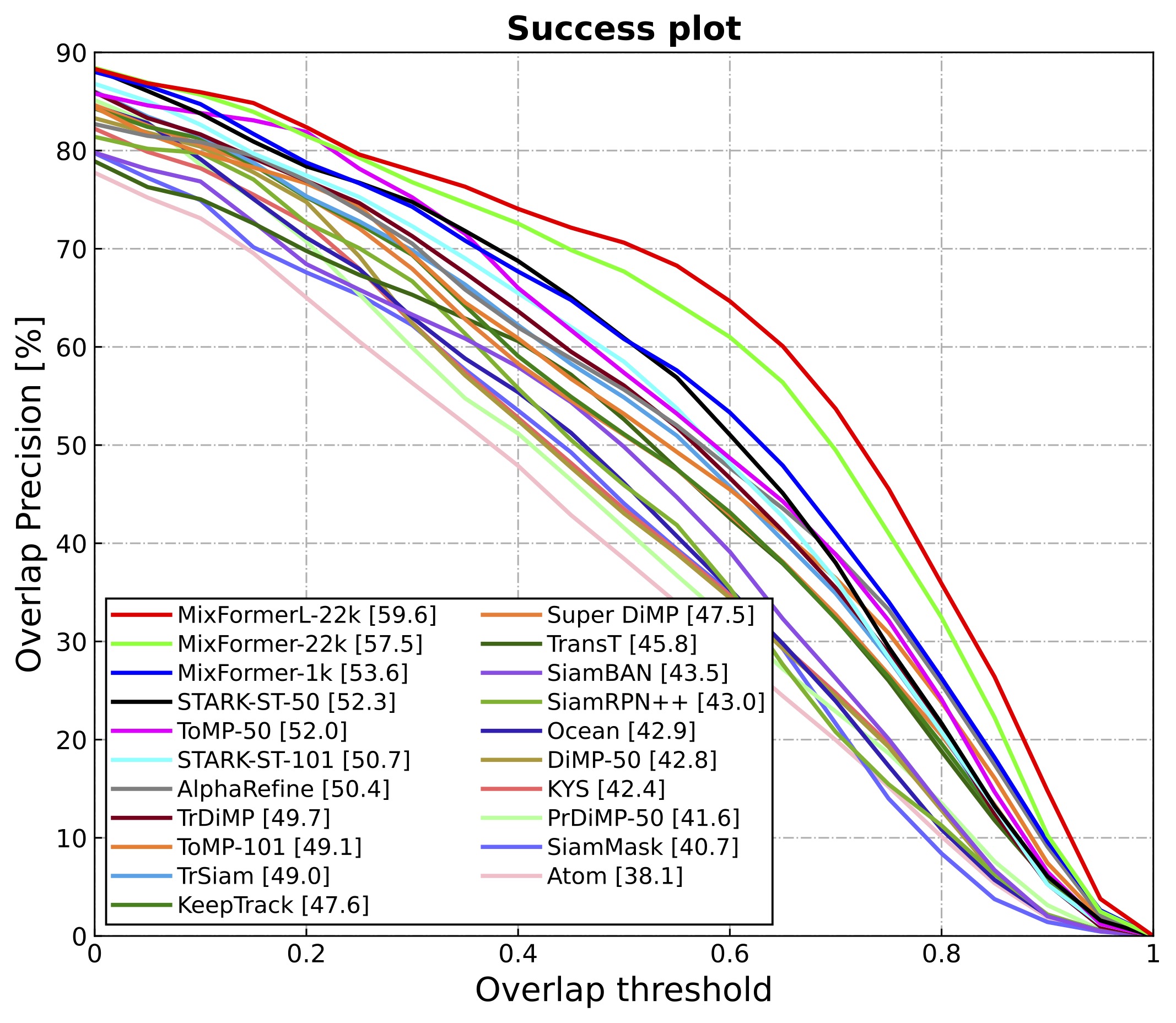}
    \vspace{-6mm}
    \caption{Camouflage}
    \label{fig:success_cm}
  \end{subfigure}%
  \end{center}
  \vspace{-1mm}
\caption{Comparisons in terms of success plots on AVisT. The AUC scores are given in the legend. In all cases, the recent MixFormer with a stronger backbone with ImageNet-22k pre-training achieves better performance. Comparing to trackers employing ImageNet-1k pre-trained backbones, ToMP-50 achieves superior overall performance. For weather conditions attribute, MixFormer with ImageNet-22k pre-training (MixFormer-22k) achieves significant improvements over its variant using ImageNet-1k backbone (MixFormer-1k). However, there is no impact of large-scale ImageNet-22k pre-training for obstruction effects with both MixFormer-22k and MixFormer-1k achieving similar results. Best viewed zoomed in.}\label{fig:success_attributes}
\end{figure}

\noindent\textbf{Siamese:} SiamRPN++~\cite{Li_2019_CVPR_SiamRPN++} employs a region proposal network to detect the target and to produce accurate bounding boxes. SiamMask~\cite{Wang_2019_CVPR_SiamMask} proposes an auxiliary binary segmentation loss and produces a segmentation mask and a bounding box for the target. SiamBAN~\cite{Chen_2020_CVPR_SiamBAN} employs a box adaptive network that fuses multi-scale features to robustly localize targets of various scales. Ocean~\cite{Zhang_2020_ECCV_Ocean} employs an object aware anchor free network for target classification and bounding box regression.   \\
\noindent\textbf{Discriminative Classifiers:} Atom~\cite{Danelljan_2019_CVPR_ATOM} uses an online trained two-layer fully convolutional neural network for target classification and employs a target estimation branch based on overlap maximization.
DiMP~\cite{Bhat_2019_ICCV_DIMP} adopts the target estimation component of Atom but proposes an end-to-end learnable optimization-based model predictor that produces discriminative filter weights to localize the target. PrDiMP~\cite{Danelljan_2020_CVPR_PRDIMP} and SuperDiMP~\cite{Danelljan_2019_github_pytracking} employ a probabilistic regression formulation. KeepTrack~\cite{Mayer_2021_ICCV_KeepTrack} uses SuperDiMP as a base tracker and employs a target candidate association network on top to reliably identify the target among distractor objects. Similarly, KYS~\cite{Bhat_2020_ECCV_KYS} employs DiMP-50 as baseline tracker but propagates dense localized state vectors that encode target, background or distractor information allowing to localized the target more robustly. In contrast, AlphaRefine~\cite{Yan_2021_CVPR_AlphaRefine} refines the preliminary bounding box generated by the base tracker SuperDiMP to increase the tracking accuracy. \\
\noindent\textbf{Transformers:} TrDiMP~\cite{Wang_2021_CVPR_TrDiMP} and TrSiam~\cite{Wang_2021_CVPR_TrDiMP} employ a transformer to enhance the extracted template and search features that are then used in a discriminative classification or Siamese setting. In contrast, TransT~\cite{Chen_2021_CVPR_TransT} directly extracts discriminative features using a feature fusion module performing self and cross attention operations. STARK~\cite{Yan_2021_ICCV_STARK} jointly fuses template and search features using a transformer encoder consisting of self attention operations and uses a transformer decoder together with an object query to predict the target state. ToMP~\cite{Mayer_2022_CVPR_Tomp} is inspired by DiMP but replaces the online-optimization based model predictor with a transformer that predicts discriminative filter weights. In contrast to aforementioned trackers that use a feature extractor followed by a feature fusion module, MixFormer~\cite{Cui_2022_CVPR_Mixformer} only uses a mixed attention based backbone that allows to directly extract discriminative features.

\subsection{Evaluation Results}
\noindent\textbf{Evaluation Metric:} Following LaSOT~\cite{Fan_2019_CVPR_Lasot}, we evaluate the tracking performance using the one-pass evaluation~\cite{WU_2015_TPAMI_OTB},  assessing the success score of different tracking methods. Success is calculated as the intersection over union (IoU) of the ground truth bounding box and the tracking result. The Area Under the Curve (AUC), which ranges from 0 to 1, is used to rank the trackers. Additionally, we present the results in terms of normalized precision plot in the suppl. material. We normalize the precision as in \cite{muller2018trackingnet}. The precision is calculated by comparing the pixel distance between the ground-truth bounding box and the tracking result. 

\noindent\textbf{Quantitative Results:}
We perform comprehensive evaluations for each of the 120 videos in AVisT. Each tracker is evaluated with publicly available trained weights. Tab.~\ref{tab:trackers} shows the performance, in terms of AUC score, of trackers for the different frameworks. Among the trackers belonging to the Siamese-based framework, SiamRPN++~\cite{Li_2019_CVPR_SiamRPN++}, SiamBAN~\cite{Chen_2020_CVPR_SiamBAN} and Ocean~\cite{Zhang_2020_ECCV_Ocean} achieve AUC scores of 39.01, 37.58 and 38.89, respectively. Within the discriminative classifier framework, KeepTrack~\cite{Mayer_2021_ICCV_KeepTrack} and AlphaRefine~\cite{Yan_2021_CVPR_AlphaRefine} achieve comparable performance with AUC score of 49.44 and 49.63, respectively. Among existing transformer-based methods employing backbones pre-trained on ImageNet-1k, STARK-ST-50 \cite{Yan_2021_ICCV_STARK} and ToMP-50 \cite{Mayer_2022_CVPR_Tomp} achieve obtain similar AUC scores of 51.11 and 51.60. The recently introduced MixFormer-1k \cite{Cui_2022_CVPR_Mixformer} achieves AUC score of 50.83.  A significant improvement in tracking performance is obtained when using ImageNet-22k pre-trained backbone in MixFormer~\cite{Cui_2022_CVPR_Mixformer}, with MixFormerL-22k achieving AUC score of 56.0. 
We also report the overall success plot in Fig.~\ref{fig:success_attributes}(a). Additional details and results are presented in suppl. material. \newline
\noindent\textbf{Attribute-based Comparison: } 
 In Fig.~\ref{fig:success_attributes} (b-f), we further evaluate the trackers on the five attributes in AVisT. We observe that a stronger backbone along with large-scale ImageNet-22k pre-training typically helps achieve better results (MixFormerL-22k~\cite{Cui_2022_CVPR_Mixformer}). However, the performance varies among attributes when using a tracker with same backbone that is either pre-trained on ImageNet-22k or ImageNet-1k (MixFormer-1k and MixFormer-22k). Further, the results also vary among attributes when using trackers all employing ImageNet-1k pre-trained backbones. For instance, MixFormer~\cite{Cui_2022_CVPR_Mixformer} significantly improves when using a backbone with ImageNet-22k pre-training on weather conditions attribute, compared to the same tracker and backbone but with ImageNet-1k pre-training (MixFormer-1k: 54.4 vs. MixFormer-22k: 58.4). However, we observe no improvement in performance possibly due to this large-scale pre-training when moving from MixFormer-1k to MixFormer-22k for obstruction effects. In case of imaging effects, ToMP-101~\cite{Mayer_2022_CVPR_Tomp} with ImageNet-1k pre-trained backbone achieves AUC score of 46.1, which is even comparable to that of MixFormerL-22k using a stronger backbone along with ImageNet-22k pre-training. For target effects, we observe MixFormer-1k to struggle compared to ToMP~\cite{Mayer_2022_CVPR_Tomp}, KeepTrack~\cite{Mayer_2021_ICCV_KeepTrack} and AlphaRefine~\cite{Yan_2021_CVPR_AlphaRefine}. To summarize, we observe that among recent trackers utilizing ImageNet-1k pre-trained backbones, no single method achieves better performance against its counterparts on all attributes. The comparisons further highlight the scope in designing a novel tracking mechanism which could tackle the diverse range of scenarios and attributes comprising sequences captured in real-world adverse conditions. More results are in suppl. material.

\section{Conclusions} We introduce a new benchmark, AVisT, for visual tracking in diverse scenarios with adverse visibility. AVisT comprises 120 challenging videos, covering 18 diverse scenarios with 42 classes. These diverse scenarios are further grouped into five attributes. We evaluate a variety of recent Siamese, discriminative classifiers and transformer-based trackers. Our experiments show that even the most recent transformer-based tracker using a heavy ImageNet-22k backbone achieves an AUC score of only 56.0\%, thereby highlighting the challenging nature of AVisT. We further analyze trackers based on attributes observing the need to design novel solutions that achieve favorable performance on real-world adverse tracking conditions. 

\bibliography{egbib}

\begin{thebibliography}{27}
\providecommand{\natexlab}[1]{#1}
\providecommand{\url}[1]{\texttt{#1}}
\expandafter\ifx\csname urlstyle\endcsname\relax
  \providecommand{\doi}[1]{doi: #1}\else
  \providecommand{\doi}{doi: \begingroup \urlstyle{rm}\Url}\fi

\bibitem[Bhat et~al.(2019)Bhat, Danelljan, Gool, and
  Timofte]{Bhat_2019_ICCV_DIMP}
Goutam Bhat, Martin Danelljan, Luc~Van Gool, and Radu Timofte.
\newblock Learning discriminative model prediction for tracking.
\newblock In \emph{Proceedings of the IEEE/CVF International Conference on
  Computer Vision (ICCV)}, October 2019.

\bibitem[Bhat et~al.(2020)Bhat, Danelljan, Van~Gool, and
  Timofte]{Bhat_2020_ECCV_KYS}
Goutam Bhat, Martin Danelljan, Luc Van~Gool, and Radu Timofte.
\newblock Know your surroundings: Exploiting scene information for object
  tracking.
\newblock In \emph{Proceedings of the European Conference on Computer Vision
  (ECCV)}, August 2020.

\bibitem[Chen et~al.(2021)Chen, Yan, Zhu, Wang, Yang, and
  Lu]{Chen_2021_CVPR_TransT}
Xin Chen, Bin Yan, Jiawen Zhu, Dong Wang, Xiaoyun Yang, and Huchuan Lu.
\newblock Transformer tracking.
\newblock In \emph{Proceedings of the IEEE/CVF Conference on Computer Vision
  and Pattern Recognition (CVPR)}, June 2021.

\bibitem[Chen et~al.(2020)Chen, Zhong, Li, Zhang, and
  Ji]{Chen_2020_CVPR_SiamBAN}
Zedu Chen, Bineng Zhong, Guorong Li, Shengping Zhang, and Rongrong Ji.
\newblock Siamese box adaptive network for visual tracking.
\newblock In \emph{Proceedings of the IEEE/CVF Conference on Computer Vision
  and Pattern Recognition (CVPR)}, June 2020.

\bibitem[Cui et~al.(2022)Cui, Jiang, Wang, and Wu]{Cui_2022_CVPR_Mixformer}
Yutao Cui, Cheng Jiang, Limin Wang, and Gangshan Wu.
\newblock Mixformer: End-to-end tracking with iterative mixed attention.
\newblock In \emph{Proceedings of the IEEE/CVF Conference on Computer Vision
  and Pattern Recognition (CVPR)}, pages 13608--13618, June 2022.

\bibitem[Danelljan and Bhat(2019)]{Danelljan_2019_github_pytracking}
Martin Danelljan and Goutam Bhat.
\newblock {PyTracking: Visual tracking library based on PyTorch.}
\newblock \url{https://github.com/visionml/pytracking}, 2019.
\newblock Accessed: 1/05/2021.

\bibitem[Danelljan et~al.(2019)Danelljan, Bhat, Khan, and
  Felsberg]{Danelljan_2019_CVPR_ATOM}
Martin Danelljan, Goutam Bhat, Fahad~Shahbaz Khan, and Michael Felsberg.
\newblock {ATOM}: Accurate tracking by overlap maximization.
\newblock In \emph{Proceedings of the IEEE/CVF Conference on Computer Vision
  and Pattern Recognition (CVPR)}, June 2019.

\bibitem[Danelljan et~al.(2020)Danelljan, Van~Gool, and
  Timofte]{Danelljan_2020_CVPR_PRDIMP}
Martin Danelljan, Luc Van~Gool, and Radu Timofte.
\newblock Probabilistic regression for visual tracking.
\newblock In \emph{Proceedings of the IEEE/CVF Conference on Computer Vision
  and Pattern Recognition (CVPR)}, June 2020.

\bibitem[Fan et~al.(2019)Fan, Lin, Yang, Chu, Deng, Yu, Bai, Xu, Liao, and
  Ling]{Fan_2019_CVPR_Lasot}
Heng Fan, Liting Lin, Fan Yang, Peng Chu, Ge~Deng, Sijia Yu, Hexin Bai, Yong
  Xu, Chunyuan Liao, and Haibin Ling.
\newblock Lasot: A high-quality benchmark for large-scale single object
  tracking.
\newblock In \emph{Proceedings of the IEEE/CVF Conference on Computer Vision
  and Pattern Recognition (CVPR)}, June 2019.

\bibitem[Galoogahi et~al.(2017)Galoogahi, Fagg, Huang, Ramanan, and
  Lucey]{Galoogahi_2017_ICCV_NFS}
Hamed~Kiani Galoogahi, Ashton Fagg, Chen Huang, Deva Ramanan, and Simon Lucey.
\newblock Need for speed: A benchmark for higher frame rate object tracking.
\newblock In \emph{ICCV}, 2017.

\bibitem[Huang et~al.(2021)Huang, Zhao, and Huang]{Huang_2021_TPAMI_GOT10k}
Lianghua Huang, Xin Zhao, and Kaiqi Huang.
\newblock Got-10k: A large high-diversity benchmark for generic object tracking
  in the wild.
\newblock \emph{IEEE Transactions on Pattern Analysis and Machine Intelligence
  (TPAMI)}, 43\penalty0 (5):\penalty0 1562--1577, 2021.

\bibitem[Kristan et~al.(2016)Kristan, Matas, Leonardis, Vojir, Pflugfelder,
  Fernandez, Nebehay, Porikli, and \v{C}ehovin]{Kristan_2016_TPAMI_VOT}
Matej Kristan, Jiri Matas, Ale\v{s} Leonardis, Tomas Vojir, Roman Pflugfelder,
  Gustavo Fernandez, Georg Nebehay, Fatih Porikli, and Luka \v{C}ehovin.
\newblock A novel performance evaluation methodology for single-target
  trackers.
\newblock \emph{IEEE Transactions on Pattern Analysis and Machine Intelligence
  (TPAMI)}, 38\penalty0 (11):\penalty0 2137--2155, 2016.

\bibitem[Lamdouar et~al.(2020)Lamdouar, Yang, Xie, and Zisserman]{MoCA}
Hala Lamdouar, Charig Yang, Weidi Xie, and Andrew Zisserman.
\newblock Betrayed by motion: Camouflaged object discovery via motion
  segmentation.
\newblock In \emph{Proceedings of the Asian Conference on Computer Vision
  (ACCV)}, November 2020.

\bibitem[Li et~al.(2019)Li, Wu, Wang, Zhang, Xing, and
  Yan]{Li_2019_CVPR_SiamRPN++}
Bo~Li, Wei Wu, Qiang Wang, Fangyi Zhang, Junliang Xing, and Junjie Yan.
\newblock Siamrpn++: Evolution of siamese visual tracking with very deep
  networks.
\newblock In \emph{Proceedings of the IEEE/CVF Conference on Computer Vision
  and Pattern Recognition (CVPR)}, June 2019.

\bibitem[Liang et~al.(2015)Liang, Blasch, and Ling]{TempleColor}
Pengpeng Liang, Erik Blasch, and Haibin Ling.
\newblock Encoding color information for visual tracking: Algorithms and
  benchmark.
\newblock \emph{IEEE Transactions on Image Processing}, 24\penalty0
  (12):\penalty0 5630--5644, 2015.
\newblock \doi{10.1109/TIP.2015.2482905}.

\bibitem[Mayer et~al.(2021)Mayer, Danelljan, Paudel, and
  Van~Gool]{Mayer_2021_ICCV_KeepTrack}
Christoph Mayer, Martin Danelljan, Danda~Pani Paudel, and Luc Van~Gool.
\newblock Learning target candidate association to keep track of what not to
  track.
\newblock In \emph{Proceedings of the IEEE/CVF International Conference on
  Computer Vision (ICCV)}, pages 13444--13454, October 2021.

\bibitem[Mayer et~al.(2022)Mayer, Danelljan, Bhat, Paul, Paudel, Yu, and
  Van~Gool]{Mayer_2022_CVPR_Tomp}
Christoph Mayer, Martin Danelljan, Goutam Bhat, Matthieu Paul, Danda~Pani
  Paudel, Fisher Yu, and Luc Van~Gool.
\newblock Transforming model prediction for tracking.
\newblock In \emph{Proceedings of the IEEE/CVF Conference on Computer Vision
  and Pattern Recognition (CVPR)}, pages 8731--8740, June 2022.

\bibitem[Mueller et~al.(2016)Mueller, Smith, and
  Ghanem]{Mueller_2016_ECCV_UAV123}
Matthias Mueller, Neil Smith, and Bernard Ghanem.
\newblock A benchmark and simulator for uav tracking.
\newblock In \emph{Proceedings of the European Conference on Computer Vision
  (ECCV)}, October 2016.

\bibitem[M{\"{u}}ller et~al.(2018)M{\"{u}}ller, Bibi, Giancola, Al{-}Subaihi,
  and Ghanem]{2018_Muller_Trackingnet}
Matthias M{\"{u}}ller, Adel Bibi, Silvio Giancola, Salman Al{-}Subaihi, and
  Bernard Ghanem.
\newblock Trackingnet: {A} large-scale dataset and benchmark for object
  tracking in the wild.
\newblock In \emph{Proceedings of the European Conference on Computer Vision
  (ECCV)}, 2018.

\bibitem[Muller et~al.(2018)Muller, Bibi, Giancola, Alsubaihi, and
  Ghanem]{muller2018trackingnet}
Matthias Muller, Adel Bibi, Silvio Giancola, Salman Alsubaihi, and Bernard
  Ghanem.
\newblock Trackingnet: A large-scale dataset and benchmark for object tracking
  in the wild.
\newblock In \emph{Proceedings of the European conference on computer vision
  (ECCV)}, pages 300--317, 2018.

\bibitem[Wang et~al.(2021)Wang, Zhou, Wang, and Li]{Wang_2021_CVPR_TrDiMP}
Ning Wang, Wengang Zhou, Jie Wang, and Houqiang Li.
\newblock Transformer meets tracker: Exploiting temporal context for robust
  visual tracking.
\newblock In \emph{Proceedings of the IEEE/CVF Conference on Computer Vision
  and Pattern Recognition (CVPR)}, June 2021.

\bibitem[Wang et~al.(2019)Wang, Zhang, Bertinetto, Hu, and
  Torr]{Wang_2019_CVPR_SiamMask}
Qiang Wang, Li~Zhang, Luca Bertinetto, Weiming Hu, and Philip~H.S. Torr.
\newblock Fast online object tracking and segmentation: A unifying approach.
\newblock In \emph{Proceedings of the IEEE/CVF Conference on Computer Vision
  and Pattern Recognition (CVPR)}, June 2019.

\bibitem[Wu et~al.(2013)Wu, Lim, and Yang]{OTB13Ming}
Yi~Wu, Jongwoo Lim, and Ming-Hsuan Yang.
\newblock Online object tracking: A benchmark.
\newblock In \emph{Proceedings of the IEEE/CVF Conference on Computer Vision
  and Pattern Recognition (CVPR)}, June 2013.

\bibitem[Wu et~al.(2015)Wu, Lim, and Yang]{WU_2015_TPAMI_OTB}
Yi~Wu, Jongwoo Lim, and Ming-Hsuan Yang.
\newblock Object tracking benchmark.
\newblock \emph{IEEE Transactions on Pattern Analysis and Machine Intelligence
  (TPAMI)}, 37\penalty0 (9):\penalty0 1834--1848, 2015.

\bibitem[Yan et~al.(2021{\natexlab{a}})Yan, Peng, Fu, Wang, and
  Lu]{Yan_2021_ICCV_STARK}
Bin Yan, Houwen Peng, Jianlong Fu, Dong Wang, and Huchuan Lu.
\newblock Learning spatio-temporal transformer for visual tracking.
\newblock In \emph{Proceedings of the IEEE/CVF International Conference on
  Computer Vision (ICCV)}, pages 10448--10457, October 2021{\natexlab{a}}.

\bibitem[Yan et~al.(2021{\natexlab{b}})Yan, Zhang, Wang, Lu, and
  Yang]{Yan_2021_CVPR_AlphaRefine}
Bin Yan, Xinyu Zhang, Dong Wang, Huchuan Lu, and Xiaoyun Yang.
\newblock Alpha-refine: Boosting tracking performance by precise bounding box
  estimation.
\newblock In \emph{Proceedings of the IEEE/CVF Conference on Computer Vision
  and Pattern Recognition (CVPR)}, June 2021{\natexlab{b}}.

\bibitem[Zhang et~al.(2020)Zhang, Peng, Fu, Li, and Hu]{Zhang_2020_ECCV_Ocean}
Zhipeng Zhang, Houwen Peng, Jianlong Fu, Bing Li, and Weiming Hu.
\newblock Ocean: Object-aware anchor-free tracking.
\newblock In \emph{Proceedings of the European Conference on Computer Vision
  (ECCV)}, August 2020.

\end{thebibliography}
\end{document}